\documentclass[runningheads]{llncs}

\usepackage{eccv}

\usepackage{eccvabbrv}

\usepackage{graphicx}
\usepackage{booktabs}

\usepackage[accsupp]{axessibility}  % Improves PDF readability for those with disabilities.

\usepackage{hyperref}
\usepackage{amsmath}
\usepackage{booktabs}
\usepackage{siunitx}
\usepackage{xcolor}
\usepackage{hyperref}
\usepackage{url}
\usepackage{xcolor}
\usepackage{array}
\definecolor{myteal}{HTML}{008080}
\newcolumntype{T}{>{\color{myteal}\arraybackslash}c}
\usepackage{orcidlink}

\begin{document}

\title{Projection-Free Transformers via Gaussian Kernel Attention}

\titlerunning{Gaussian Kernel Attention}

\author{Debarshi Kundu \and
Archisman Ghosh \and
Swaroop Ghosh
\and
Vasant Honavar}
\authorrunning{D.Kundu et al.}

\institute{The Pennsylvania State University, University Park, USA
\email{vuh14@psu.edu}\\
}

\maketitle

% ---------------------------------------------------------------
% ABSTRACT
\begin{abstract}
Self-attention in Transformers is typically implemented as 
$\mathrm{softmax}(QK^\top/\sqrt{d})V$, where $Q=XW_Q$, $K=XW_K$, and $V=XW_V$ 
are learned linear projections of the input $X$. 
We ask whether these learned projections are necessary, or whether they can be replaced by a simpler similarity-based diffusion operator.
We introduce \textbf{Gaussian Kernel Attention} (GKA), a drop-in replacement for dot-product attention that computes token affinities directly using a Gaussian radial basis function (RBF) kernel applied to per-head token features. 
Each head learns only a bandwidth parameter $\sigma_h$, while a single output projection $W_O$ preserves compatibility with the standard Transformer interface.
GKA can be interpreted as normalized kernel regression over tokens, linking modern Transformer architectures to classical non-local filtering and kernel smoothing methods. We evaluate GKA in both vision and language modeling settings. 
For autoregressive language modeling within the \texttt{nanochat} framework, we implement causal masking and sliding-window constraints by masking and renormalizing the Gaussian kernel. 
At depth 20, a GKA model with $0.42\times$ the parameters and $0.49\times$ the total training FLOPs of a standard attention baseline trains stably, exhibits a near-zero train-validation gap, and demonstrates competitive behavior on standard benchmarks, albeit with higher bits-per-byte (BPB) at this compute scale. Overall, GKA provides a minimal, interpretable attention mechanism with an explicit locality scale, offering a dimension in the accuracy-efficiency trade-off for Transformer design.

\keywords{Vision Transformer \and Language Model \and Self-Attention \and Image Classification \and Kernelized Attention \and Projection-Free Transformers}

\end{abstract}

\section{Introduction}

\begin{figure*}[t]
    \centering
    \includegraphics[width=\linewidth]{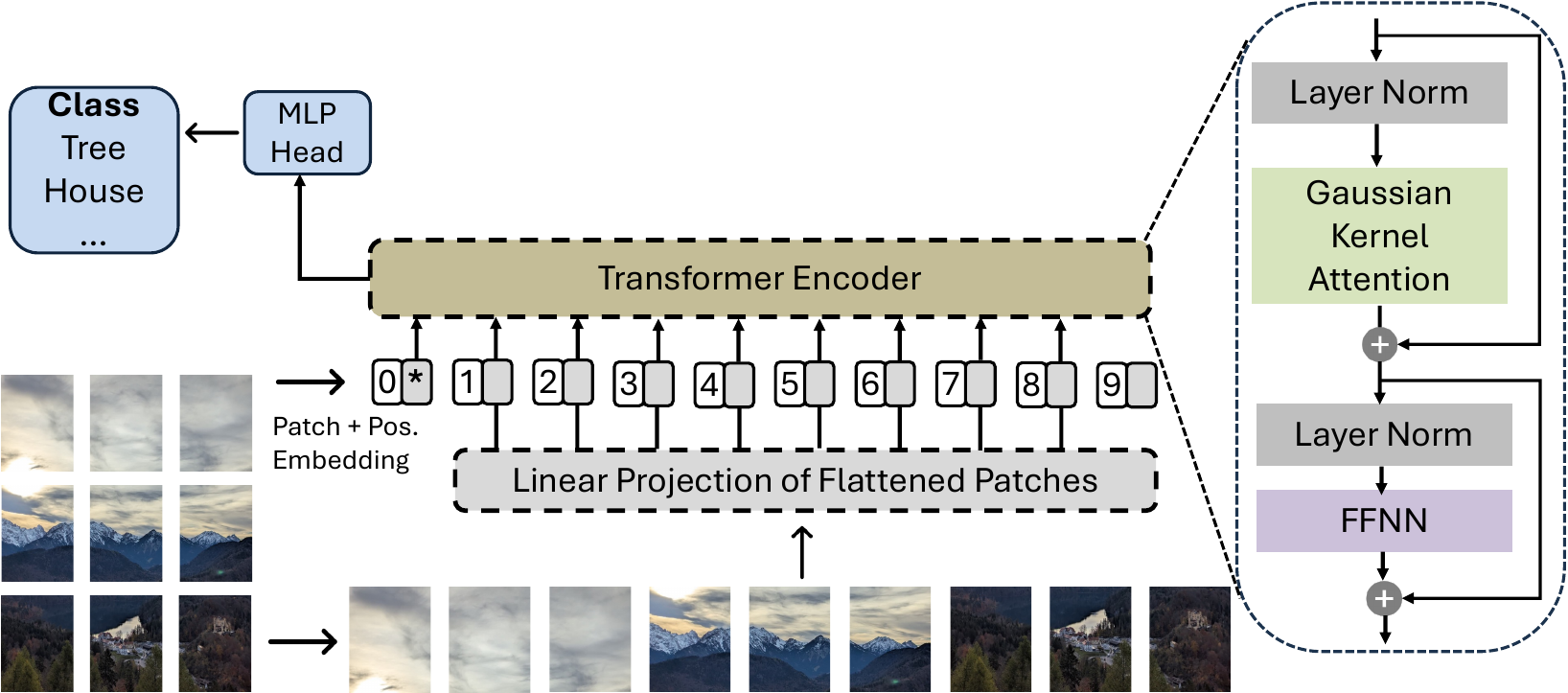}
    \caption{\textbf{\textsc{GKA-Transformer}.} Projection-based self-attention is replaced with Gaussian Kernel Attention (see \cref{fig:computation}).}
\label{fig:block}
\end{figure*}

Vision Transformers (ViTs)~\cite{dosovitskiy2021image} established global self-attention as a competitive alternative to convolution for large-scale image recognition. Subsequent refinements, like data-efficient training~\cite{touvron2021training}, hierarchical designs~\cite{liu2021swin}, and improved scaling, retain the same projection-based dot-product attention introduced in~\cite{vaswani2017attention}. Despite extensive architectural innovation, the attention operator itself remains largely unchanged.
In a standard attention block,
\[
\mathrm{Attn}(X)=\mathrm{softmax}(QK^\top/\sqrt{d})V,\qquad
Q=XW_Q,\;K=XW_K,\;V=XW_V,
\]
the learned projections $W_Q,W_K,W_V$ define the similarity geometry and account for a substantial fraction of parameters and FLOPs. While prior work has proposed low-rank, kernelized, or linear approximations of softmax attention~\cite{katharopoulos2020transformers,choromanski2021performer,wang2020linformer,xiong2021nystromformer,chen2021skyformer,luo2021kernelizedrpe}, these methods retain learned projections.
We revisit this design choice and remove them entirely.

\paragraph{\textbf{Gaussian Kernel Attention.}}
We introduce \textbf{Gaussian Kernel Attention (GKA)} (Fig. \ref{fig:block}), a projection-free attention operator that replaces query–key dot products with Gaussian kernel diffusion over token features. For each head $h$, affinities are computed using a Gaussian RBF kernel with learnable bandwidth $\sigma_h$:
\[
K^{(h)}_{ij}=\exp\!\left(-\frac{\|x_i^{(h)}-x_j^{(h)}\|_2^2}{2\sigma_h^2}\right),\qquad
W^{(h)}=\mathrm{row\_norm}(K^{(h)}),
\]
followed by aggregation $y_i^{(h)}=\sum_j W^{(h)}_{ij}x_j^{(h)}$ and a standard output projection $W_O$.

Each head learns only a single scalar controlling locality. GKA eliminates $W_Q,W_K,W_V$ and decouples token mixing from learned projection geometry. Unlike kernelized approximations of softmax attention, GKA replaces projection-based similarity altogether with symmetric, normalized kernel diffusion. The formulation connects attention to classical non-local filtering~\cite{buades2005nonlocal,wang2018nonlocal} while remaining fully compatible with standard Transformer blocks.

\paragraph{\textbf{ImageNet evaluation.}}
We integrate GKA into DeiT-style ViTs and evaluate on ImageNet classification. Across model scales, GKA substantially reduces attention parameters and total training FLOPs while maintaining competitive top-1 accuracy. Training remains stable, and learned bandwidths specialize across heads, revealing interpretable locality patterns. These results indicate that learned query–key geometry is not strictly required for effective global patch interaction and define a distinct accuracy–efficiency trade-off frontier.

\paragraph{\textbf{Beyond vision.}}
We further evaluate GKA in a GPT-style autoregressive language model using the \texttt{nanochat} framework~\cite{nanochat_repo}, implementing causal masking and sliding-window constraints via masked kernel renormalization. In a depth-20 configuration (sequence length 2{,}048), GKA reduces parameters and total training FLOPs to $0.49\times$ that of a standard baseline while training stably. Although bits-per-byte is higher at this scale, downstream DCLM-aligned evaluation shows competitive and task-dependent behavior, indicating that projection-free kernel diffusion remains viable under autoregressive constraints.

\paragraph{\textbf{Contributions.}} The key contributions of the paper are:
\begin{itemize}
  \item \textbf{Projection-free attention.} We introduce Gaussian Kernel Attention (GKA), removing $W_Q,W_K,W_V$ and replacing projection-based similarity with Gaussian kernel diffusion using only per-head bandwidth parameters.
  \item \textbf{Competitive ViTs at reduced cost.} GKA trains stably in DeiT-style ViTs and achieves competitive ImageNet accuracy while substantially reducing attention parameters and training FLOPs.
  \item \textbf{Cross-modal validation.} GKA remains trainable and competitive in causal language modeling under autoregressive masking.
  \item \textbf{A new accuracy–efficiency frontier.} By decoupling token mixing from learned projection geometry, GKA exposes an explicit locality scale per head and defines a distinct point in the Transformer design space.
\end{itemize}

\section{Related Work}

\begin{table}[]

\caption{Compact comparison of related attention mechanisms and approximations.}
\label{tab:attn_compare}
\begin{tabular}{l|l|l|l|l}
Method              & Kernel Type & \begin{tabular}[c]{@{}l@{}}QKV \\ Projection\end{tabular} & Domain              & Notes                                                                                   \\ \hline \hline
\begin{tabular}[c]{@{}l@{}}Standard \\ Attention\cite{vaswani2017attention}\end{tabular} & \begin{tabular}[c]{@{}l@{}}Softmax dot \\ Product\end{tabular}                                    & Yes                                                       & Vision/NLP          & Baseline Attention                                                                      \\ \hline
\begin{tabular}[c]{@{}l@{}}Non-Local \\ Means\cite{buades2005nonlocal}\end{tabular}     & Gaussian-like                                          & N/A                                                       & Vision              & \begin{tabular}[c]{@{}l@{}}Similarity-weighted \\ Averaging\end{tabular}                \\ \hline
Non-Local NN\cite{wang2018nonlocal}       & Learned Affinity                                       & Partial                                                   & Vision/Video        & Non-local Block                                                                         \\ \hline
\begin{tabular}[c]{@{}l@{}}Linear \\ Transformer\cite{katharopoulos2020transformers}\end{tabular}  & \begin{tabular}[c]{@{}l@{}}Feature-Map \\ Kernel\end{tabular}                                     & Yes                                                       & NLP                 & \begin{tabular}[c]{@{}l@{}}Kernel Feature \\ Maps\end{tabular}                                                                     \\ \hline
Performer\cite{choromanski2021performer}           & \begin{tabular}[c]{@{}l@{}}Softmax \\ Approx (RFF)\end{tabular}                                   & Yes                                                       & Long-Sequence       & FAVOR+features                                                                          \\ \hline
Linformer\cite{wang2020linformer}          & Low-Rank Approx                                        & Yes                                                       & NLP                 & \begin{tabular}[c]{@{}l@{}}Learned Low-Rank \\ Projections\end{tabular}                 \\ \hline
Nyströmformer\cite{xiong2021nystromformer}       & Landmark Approx                                        & Yes                                                       & NLP                 & \begin{tabular}[c]{@{}l@{}}Nyström \\ Reconstruction\end{tabular}                       \\ \hline
Kernelized RPE\cite{luo2021kernelizedrpe}      & Kernel + RPE                                           & Yes                                                       & Vision/NLP          & \begin{tabular}[c]{@{}l@{}}FFT via Toeplitz \\ RPE\end{tabular}                         \\ \hline
Skyformer\cite{chen2021skyformer}           & Gaussian Kernel                                        & Yes                                                       & NLP                 & Gaussian+Nystrom                                                                        \\ \hline
Krause\cite{liu2026krause}              & Distance+RBF                                           & Yes                                                       & Vision/NLP          & \begin{tabular}[c]{@{}l@{}}Bounded-Confidence\\ Sparsity\end{tabular}                   \\ \hline
\textbf{GKA (ours)} & \textbf{Gaussian RBF}                                  & \textbf{No}                                               & \textbf{Vision/NLP} & \textbf{\begin{tabular}[c]{@{}l@{}}Per-Head $\sigma_h$;\\ Projection-Free\end{tabular}} \\ \hline
\end{tabular}
\end{table}

\subsection{Transformers in vision and language}

Transformers \cite{vaswani2017attention} parameterize token interactions via learned $W_Q,W_K,W_V$ projections and softmax-normalized dot products.  
In vision, ViT \cite{dosovitskiy2021image} introduced patch tokens; DeiT \cite{touvron2021training} improved data-efficient ImageNet training; Swin \cite{liu2021swin} introduced hierarchical windowed attention for dense tasks.

We remain within this backbone family (patch tokens, residual MLP blocks, LayerNorm) but modify the attention primitive itself: we remove $W_Q,W_K,W_V$ and replace dot-product softmax attention with Gaussian RBF similarity computed directly on per-head features.

To test generality, we also evaluate GKA in causal language models using \texttt{nanochat} \cite{nanochat_repo}, reporting language modeling quality and DCLM/DataComp-LM metrics \cite{dclm_paper,dclm_repo} with FineWeb-Edu pretraining \cite{finewebedu}. Benchmarks include HellaSwag \cite{hellaswag}, LAMBADA \cite{lambada}, OpenBookQA \cite{openbookqa}, WinoGrande \cite{winogrande}, and COPA \cite{copa}. This dual-domain setting stresses bidirectional mixing (vision) versus causally masked mixing (LM).

\subsection{Non-local aggregation and kernel smoothing}

Long-range aggregation in vision predates Transformers.  
Non-Local Means \cite{buades2005nonlocal} replaces a pixel/patch with a similarity-weighted average; Non-Local Neural Networks \cite{wang2018nonlocal} embed this principle as a neural block computing global weighted sums.

GKA inherits this non-local motif, construct an affinity matrix from feature similarity, and apply normalized aggregation, but implements it as a Transformer-native attention replacement. Crucially, unlike non-local blocks and standard attention, GKA does \emph{not} learn separate query/key/value embeddings: it performs Gaussian smoothing directly on per-head residual features and learns only a bandwidth $\sigma_h$ (plus output projection $W_O$).
Classical kernel methods motivate scalable similarity computation. Random Features \cite{rahimi2007random} approximate shift-invariant kernels such as Gaussian RBF. While we compute exact Gaussian affinities at moderate sequence lengths, these methods inform potential scaling extensions.

\subsection{Kernelized and efficient attention}

A large literature reduces quadratic attention cost while preserving the Transformer interface.  
Linear Transformers \cite{katharopoulos2020transformers} express attention via kernel feature maps for $\mathcal{O}(n)$ complexity. Performer \cite{choromanski2021performer} approximates softmax with positive random features. Linformer \cite{wang2020linformer} and Nyströmformer \cite{xiong2021nystromformer} exploit low-rank structure. Kernelized RPE \cite{luo2021kernelizedrpe} leverages Toeplitz structure for $\mathcal{O}(n\log n)$ computation. Skyformer \cite{chen2021skyformer} replaces softmax with a Gaussian kernel combined with Nyström approximation.

These methods modify or approximate the \emph{kernel inside projection-based attention}. In contrast, GKA removes $W_Q,W_K,W_V$ entirely and computes Gaussian RBF similarity directly on per-head features. It is therefore not a softmax approximation, low-rank surrogate, or feature-map rewrite, but a projection-free diffusion operator embedded in the Transformer block.

Optimized kernels such as FlashAttention \cite{dao2022flashattention} demonstrate that wall-clock efficiency depends on IO-aware implementations; this motivates careful systems evaluation for exact quadratic operators such as exact Gaussian kernel attention.

\subsection{Distance-based and localized attention dynamics}

Recent work revisits attention dynamics induced by kernel choice. Krause Synchronization Transformers \cite{liu2026krause} replace dot-product similarity with Euclidean distance mapped through an RBF kernel and impose bounded-confidence sparsity to reduce complexity.

Krause is complementary but distinct: while both use distance-based kernels and explicit scale parameters, Krause retains projection-based structure and focuses on sparsity dynamics. GKA instead studies a projection-free Gaussian diffusion operator and evaluates it consistently across vision and causal language modeling. \\
Table \ref{tab:attn_compare} summaries various related works and situates our work. 
\section{Gaussian Kernel Attention}
\label{sec:method}

\begin{figure*}[t]
    \centering
    \includegraphics[width=01\linewidth]{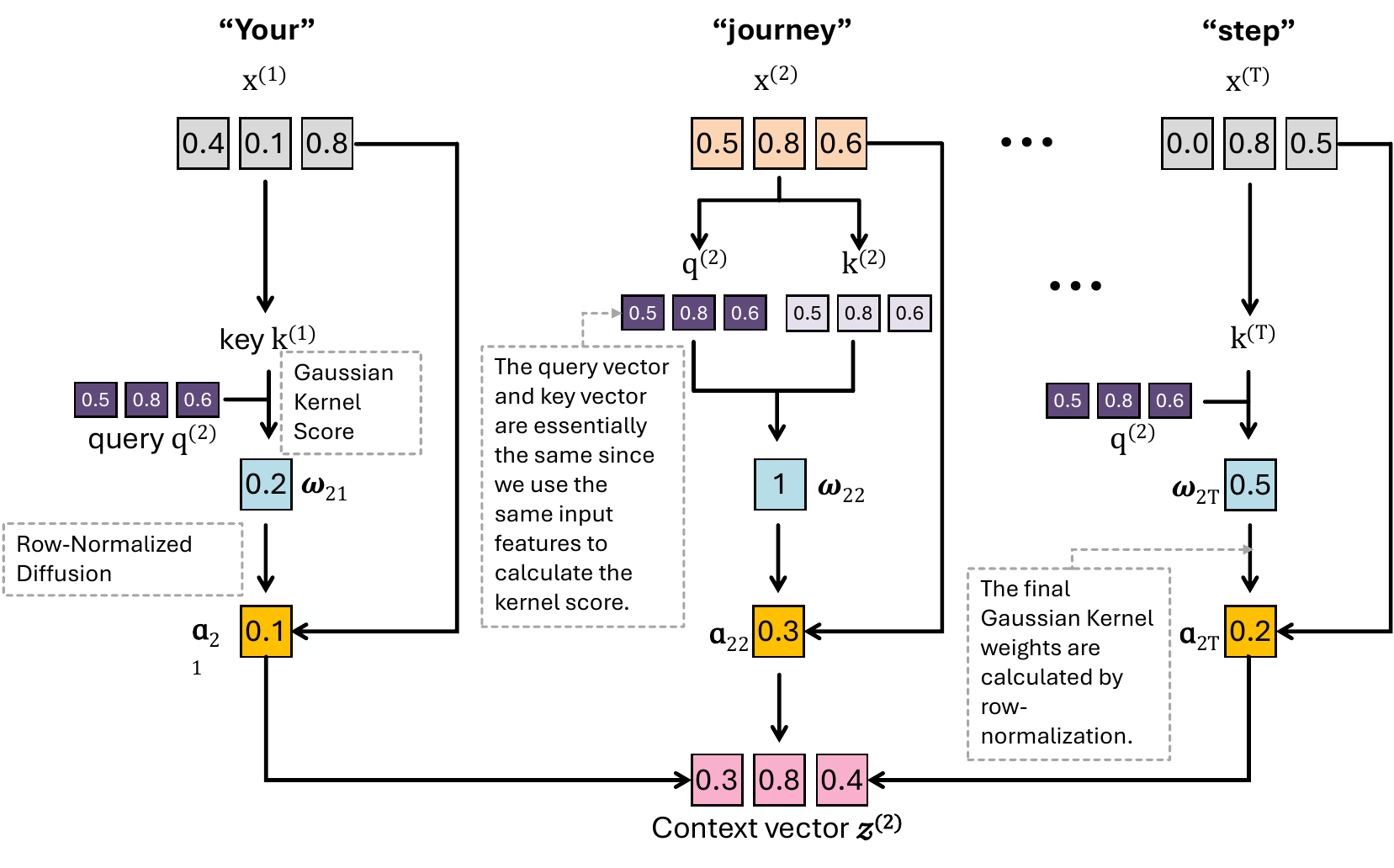}
\caption{\textbf{Gaussian Kernel Attention (simplified illustrative example).}
For token $i$, we compute Gaussian similarities to all tokens, row-normalize the affinities, and form the output as a weighted sum. Feature-similar tokens contribute more, yielding smooth normalized non-local aggregation.}
\label{fig:computation}
\end{figure*}

\subsection{Preliminaries}

Let $X \in \mathbb{R}^{B \times N \times D}$ denote token sequences (vision patches or text), where $B$ is the batch size, $N$ tokens, and $D$ model width. A Transformer block alternates token mixing with an MLP under residual connections and normalization \cite{vaswani2017attention}. Standard multi-head attention computes
\begin{equation}
\mathrm{Attn}(X) =
\mathrm{Concat}_{h=1}^H
\big(\mathrm{softmax}(Q^{(h)} {K^{(h)}}^\top / \sqrt{d}) V^{(h)}\big)
W_O,
\end{equation}
with $d=D/H$ and $Q=XW_Q$, $K=XW_K$, $V=XW_V$.

We retain this block interface while replacing projection-based softmax attention with projection-free kernel diffusion.

\subsection{Gaussian Kernel Attention}
\label{sec:gka}
The illustration in Fig. \ref{fig:computation} shows the GKA computation for a single head.
Given normalized tokens $\tilde{X}$ (e.g., pre-attention LayerNorm output), we reshape into heads:
\begin{equation}
\tilde{X} \mapsto \tilde{X}^{(h)} \in \mathbb{R}^{B \times N \times d}, \quad h=1,\dots,H.
\end{equation}

For each head, we define a Gaussian RBF affinity:
\begin{equation}
K^{(h)}_{ij}
= \exp\!\left(-\frac{\|\tilde{x}^{(h)}_i - \tilde{x}^{(h)}_j\|_2^2}{2\sigma_h^2}\right),
\label{eq:rbf_kernel}
\end{equation}
with learnable bandwidth $\sigma_h = \exp(\ell_h)$.

\paragraph{Row-stochastic diffusion.}
Affinities are converted into a row-stochastic mixing matrix:
\begin{equation}
W^{(h)}_{ij}
= \frac{K^{(h)}_{ij}}{\sum_{j'} K^{(h)}_{ij'} + \varepsilon},
\qquad
Y^{(h)}_i
= \sum_{j=1}^{N} W^{(h)}_{ij}\,\tilde{x}^{(h)}_j.
\label{eq:row_norm}
\end{equation}

Each head thus applies Gaussian kernel smoothing—equivalently, one step of feature-space diffusion on a fully connected similarity graph. No $W_Q,W_K,W_V$ are learned; only $\{\ell_h\}_{h=1}^H$ and $W_O$ remain.

\paragraph{\textbf{Output projection}.}
\begin{equation}
\mathrm{GKA}(\tilde{X}) =
\mathrm{Concat}_{h=1}^H(Y^{(h)})\, W_O.
\end{equation}

\paragraph{\textbf{Efficient distance computation.}}
Squared distances use
\begin{equation}
\|a-b\|_2^2 = \|a\|_2^2 + \|b\|_2^2 - 2a^\top b,
\end{equation}
so the dominant cost reduces to batched matrix multiplications followed by exponentiation and normalization. 

\subsection{Masking}
\label{sec:masking}

Masking is applied directly to affinities before normalization. Let $\mathbb{I}[\mathrm{allowed}(i,j)]$ be a binary mask:
\begin{equation}
\bar{K}^{(h)}_{ij} = K^{(h)}_{ij}\,\mathbb{I}[\mathrm{allowed}(i,j)],
\qquad
W^{(h)}_{ij}
= \frac{\bar{K}^{(h)}_{ij}}{\sum_{j'} \bar{K}^{(h)}_{ij'} + \varepsilon}.
\label{eq:masked_renorm}
\end{equation}

For language models, causal masking enforces $j \le i$.  
Optional sliding-window masking additionally restricts $i-W < j \le i$, reducing effective complexity to linear in $N$ up to window size.

\subsection{Integration into Transformer blocks}
\label{sec:integration}

GKA replaces multi-head attention in a standard pre-norm block:
\begin{align}
\tilde{X} &= \mathrm{LN}(X),\\
X' &= X + \mathrm{DropPath}(\mathrm{GKA}(\tilde{X})),\\
Y &= X' + \mathrm{DropPath}(\mathrm{MLP}(\mathrm{LN}(X'))).
\end{align}

Vision models apply bidirectional mixing over \texttt{[CLS]} and patch tokens; language models apply causal (optionally windowed) masking.

\subsection{Complexity}
\label{sec:complexity}

Unmasked GKA computes an $N \times N$ affinity per head, giving time $\mathcal{O}(B H N^2 d)$ and memory $\mathcal{O}(B H N^2)$.  
With window size $W$, complexity becomes $\mathcal{O}(B H N W d)$ time and $\mathcal{O}(B H N W)$ memory.

As an exact quadratic operator under the chosen mask, GKA benefits from IO-aware attention implementations such as FlashAttention \cite{dao2022flashattention}, since computation reduces to matrix multiplications plus pointwise operations amenable to fused kernels.

\section{Experiments in Vision Task}
\label{sec:experiments}

\subsection{Experimental Setup}
\label{sec:setup}

\paragraph{\textbf{Dataset.}}
We evaluate on ImageNet-1K~\cite{deng2009imagenet} (1.28M training, 50k validation images; 1,000 classes). All models are trained from scratch without external data or distillation.

\paragraph{\textbf{Training protocol.}}
For fair comparison, we follow the DeiT recipe~\cite{touvron2021deit} across all models. We train for 300 epochs with AdamW~\cite{loshchilov2019adamw}, base learning rate $10^{-3}$ scaled by effective batch size ($\mathrm{lr}=\mathrm{lr}_{\mathrm{base}}\cdot \mathrm{batch}_{\mathrm{eff}}/1024$), weight decay $0.05$, and $(\beta_1,\beta_2)=(0.9,0.999)$. We use 5-epoch linear warmup and cosine decay to $10^{-6}$. Weight decay is disabled for biases, LayerNorm parameters, positional embeddings, the \texttt{[CLS]} token, and the learnable $\log\sigma$ parameters.

\paragraph{\textbf{Data augmentation and regularization.}}
We use random resized crop to $224{\times}224$ (scale $[0.08,1.0]$, bicubic), random horizontal flip, RandAugment~\cite{cubuk2020randaugment} ($n{=}2$, $m{=}9$), random erasing~\cite{zhong2020random} ($p{=}0.25$), Mixup~\cite{zhang2018mixup} ($\alpha{=}0.8$), CutMix~\cite{yun2019cutmix} ($\alpha{=}1.0$), label smoothing~\cite{szegedy2016rethinking} ($\epsilon{=}0.1$), and repeated augmentation~\cite{hoffer2020augment} (3 repeats). We apply stochastic depth~\cite{huang2016stochastic} with drop rate $0.1$ for Tiny/Small and $0.2$ for Base.

\paragraph{\textbf{Optimization details and reporting.}}
We use EMA weights (decay $0.99996$) and report EMA Top-1 as the primary accuracy metric. Gradient clipping uses max norm $1.0$.

\paragraph{\textbf{Infrastructure.}}
Training uses 7$\times$ NVIDIA RTX 5090 GPUs (32\,GB) with PyTorch DDP (NCCL), BF16 mixed precision, and \texttt{torch.compile} (\texttt{max-autotune}). We use per-GPU batch size 256 (effective batch size 1,792). Throughput and memory are averaged over 100 forward passes after warmup.

\subsection{Results on the Vision Task}
\label{sec:main_results}

Table~\ref{tab:main_results} compares GKA against DeiT baselines across model scales. We report total parameters, attention-module parameters, FLOPs, checkpoint size, and ImageNet Top-1 accuracy.

% \begin{table}[t]
% \centering
% \caption{\textbf{Comparison of GKA and DeiT on ImageNet-1K.} \textbf{Attn} counts parameters within attention modules only; \textbf{MLP} parameters match by design. Top-1 uses EMA weights. $\Delta$Params and $\Delta$FLOPs are relative to the DeiT baseline at the same scale.}
% \label{tab:main_results}
% \vspace{2mm}
% \resizebox{\textwidth}{!}{%
% \begin{tabular}{l|ccc|cc|c|cc}
% \toprule
% \textbf{Model} & \textbf{Total} & \textbf{Attn} & \textbf{MLP} & \textbf{FLOPs} & \textbf{Size} & \textbf{Top-1} & $\boldsymbol{\Delta}$\textbf{Params} & $\boldsymbol{\Delta}$\textbf{FLOPs} \\
%  & (M) & (M) & (M) & (G) & (MB) & (\%) & (\%) & (\%) \\
% \midrule
% DeiT-Ti & 5.72 & 1.78 & 3.55 & 2.51 & 21.9 & 72.2 & --- & --- \\
% GKA-Ti  & 4.38 & 0.44 & 3.55 & 1.98 & 16.8 & 71.54 & $-$23.4 & $-$21.1 \\
% \midrule
% DeiT-S  & 22.05 & 7.10 & 14.18 & 9.20 & 84.2 & 79.8 & --- & --- \\
% GKA-S   & 16.73 & 1.77 & 14.18 & 7.11 & 63.9 & 78.23 & $-$24.1 & $-$22.7 \\
% \midrule
% DeiT-B  & 86.57 & 28.35 & 56.67 & 35.13 & 330.3 & 81.8 & --- & --- \\
% GKA-B   & 65.31 & 7.09 & 56.67 & 26.76 & 249.2 & 80.3 & $-$24.6 & $-$23.8 \\
% \bottomrule
% \end{tabular}}
% \end{table}
\begin{table}[t]
\centering
\caption{\textbf{Comparison of GKA and DeiT on ImageNet-1K.} \textbf{Attn} counts parameters within attention modules only; \textbf{MLP} parameters match by design. Top-1 uses EMA weights. $\Delta$Params and $\Delta$FLOPs are relative to the DeiT baseline at the same scale.}
\label{tab:main_results}
\vspace{2mm}
\resizebox{\textwidth}{!}{%
\begin{tabular}{l|ccc|cc|c|TT}
\toprule
\textbf{Model} & \textbf{Total} & \textbf{Attn} & \textbf{MLP} & \textbf{FLOPs} & \textbf{Size} & \textbf{Top-1} & $\boldsymbol{\Delta}$\textbf{Params} & $\boldsymbol{\Delta}$\textbf{FLOPs} \\
 & (M) & (M) & (M) & (G) & (MB) & (\%) & (\%) & (\%) \\
\midrule
DeiT-Ti & 5.72 & 1.78 & 3.55 & 2.51 & 21.9 & 72.2 & --- & --- \\
\textbf{GKA-Ti}  & \textbf{4.38} & \textbf{0.44} & 3.55 & \textbf{1.98} & \textbf{16.8} & 71.54 & \textbf{$-$23.4} & \textbf{$-$21.1} \\
\midrule
DeiT-S  & 22.05 & 7.10 & 14.18 & 9.20 & 84.2 & 79.8 & --- & --- \\
\textbf{GKA-S}   & \textbf{16.73} & \textbf{1.77} & 14.18 & \textbf{7.11} & \textbf{63.9} & 78.23 & \textbf{$-$24.1 }& \textbf{$-$22.7} \\
\midrule
DeiT-B  & 86.57 & 28.35 & 56.67 & 35.13 & 330.3 & 81.8 & --- & --- \\
\textbf{GKA-B }  & \textbf{65.31} & \textbf{7.09} & 56.67 & \textbf{26.76} & \textbf{249.2} & 80.3 & \textbf{$-$24.6} & \textbf{$-$23.8} \\
\bottomrule
\end{tabular}}
\end{table}
Across scales, GKA reduces total parameters by 23--25\% and FLOPs by 21--24\%, with proportional checkpoint (Ckpt) savings (e.g., 84.2\,MB $\rightarrow$ 63.9\,MB for Small). These gains come entirely from attention: removing $W_Q,W_K,W_V$ reduces attention-module parameters by $\approx$75\%, while MLP capacity is unchanged.

Accuracy degrades modestly. GKA-Ti trails DeiT-Ti by 0.66\,pp (71.54 vs.\ 72.2), and GKA-S trails DeiT-S by 1.57\,pp (78.23 vs.\ 79.8). These results suggest that a large portion of projection-based attention capacity can be replaced by distance-based Gaussian diffusion on residual-stream features, while retaining competitive ImageNet performance at lower parameter and FLOP budgets.

\subsection{Efficiency Analysis}
\label{sec:efficiency}

The parameter/FLOP reductions above do not directly translate into wall-clock gains. We therefore measure runtime and memory explicitly. Inference is benchmarked on a single GPU over batch sizes $\{32,64,128,256,512\}$ and we report peak throughput. Training measurements use DDP across 7 GPUs in BF16 at per-GPU batch sizes 64 and 128.

\begin{table}[t]
    \centering
    \caption{\textbf{Parameter and compute efficiency.} GKA removes $W_q,W_k,W_v$ and replaces them with $H$ learnable $\sigma_h$ scalars per block. Attention parameters include $W_o$ and, for GKA, the $\sigma$ scalars. FLOPs are analytical single-image forward-pass estimates.}
    \label{tab:param_efficiency}
    \vspace{2mm}
    \resizebox{\columnwidth}{!}{%
    \begin{tabular}{l|c|c|c|c|c|c}
        \toprule
        \textbf{Model} & \textbf{Total} & \textbf{Attn} & \textbf{MLP} & \textbf{$\sigma$ Params} & \textbf{FLOPs} & \textbf{Ckpt} \\
         & (M) & (M) & (M) & & (G) & (MB) \\
        \midrule
        DeiT-Ti  & 5.72  & 1.78  & 3.55 & ---  & 2.51  & 21.9 \\
        \textbf{GKA-Ti}   & \textbf{4.38} \textcolor{teal}{\scriptsize(\textbf{$-$23\%})}  & \textbf{0.44} \textcolor{teal}{\scriptsize(\textbf{$-$75\%})}  & 3.55 & 36  & \textbf{1.98} \textcolor{teal}{\scriptsize(\textbf{$-$21\%})}  & \textbf{16.7} \\
        \midrule
        DeiT-S   & 22.05  & 7.10  & 14.18 & ---  & 9.22  & 84.2 \\
        \textbf{GKA-S }   & \textbf{16.73} \textcolor{teal}{\scriptsize(\textbf{$-$24\%})}  & \textbf{1.77} \textcolor{teal}{\scriptsize(\textbf{$-$75\%})}  & 14.18 & 72  & \textbf{7.11} \textcolor{teal}{\scriptsize(\textbf{$-$23\%})}  & \textbf{63.9} \\
        \midrule
        DeiT-B   & 86.57  & 28.35  & 56.67 & ---  & 35.13  & 330.2 \\
        \textbf{GKA-B}    & \textbf{65.31} \textcolor{teal}{\scriptsize(\textbf{$-$25\%})}  & \textbf{7.09} \textcolor{teal}{\scriptsize(\textbf{$-$75\%})}  & 56.67 & 144  & \textbf{26.76} \textcolor{teal}{\scriptsize(\textbf{$-$24\%})}  & \textbf{249.1} \\
        \bottomrule
    \end{tabular}%
    }
    \vspace*{-0.2in}
\end{table}

\begin{table}[t]
    \centering
    \caption{\textbf{Runtime efficiency on 7$\times$\,RTX~5090.} Inference throughput/memory are measured on a single GPU at the best batch size (in parentheses). Training uses DDP across 7 GPUs at the indicated per-GPU batch size (BF16). $\Delta$ is relative to DeiT at the same scale.}
    \label{tab:runtime_efficiency}
    \vspace{2mm}
    \resizebox{\columnwidth}{!}{%
    \begin{tabular}{l|c|c|c|c|c|c}
        \toprule
        & \multicolumn{2}{c}{\textbf{Inference}} & \multicolumn{2}{c}{\textbf{Training (bs\,=\,64/gpu)}} & \multicolumn{2}{c}{\textbf{Training (bs\,=\,128/gpu)}} \\
        \cmidrule(lr){2-3} \cmidrule(lr){4-5} \cmidrule(lr){6-7}
        \textbf{Model} & Throughput & Mem & Throughput & Mem & Throughput & Mem \\
         & (img/s) & (MB) & (img/s) & (MB) & (img/s) & (MB) \\
        \midrule
        DeiT-Ti  & 43{,}182\,(128) & 316  & 12{,}131 & 1{,}245   & 23{,}258 & 2{,}352 \\
        \textbf{GKA-Ti}   & 37{,}806\,(128) & 311  & 7{,}567  & 2{,}484   & 13{,}678 & 4{,}876 \\
        $\Delta$\,\%  & \textcolor{red}{$-$12\%} & $-$2\%  & \textcolor{red}{$-$38\%} & \textcolor{red}{$+$100\%} & \textcolor{red}{$-$41\%} & \textcolor{red}{$+$107\%} \\
        \midrule
        DeiT-S   & 16{,}199\,(64)  & 378  & 10{,}010 & 2{,}569   & 20{,}100 & 4{,}710 \\
        \textbf{GKA-S} & 15{,}649\,(64)  & 358  & 6{,}519  & 5{,}028   & 7{,}394  & 9{,}755 \\
        $\Delta$\,\%  & \textcolor{red}{$-$3\%}  & $-$5\%  & \textcolor{red}{$-$35\%} & \textcolor{red}{$+$96\%}  & \textcolor{red}{$-$63\%} & \textcolor{red}{$+$107\%} \\
        \midrule
        DeiT-B   & 4{,}225\,(64)   & 3{,}165  & 7{,}672  & 5{,}961   & 7{,}976  & 10{,}089 \\
        \textbf{GKA-B}    & 2{,}288\,(32)   & 4{,}753  & 3{,}815  & 10{,}528  & 3{,}385  & 19{,}918 \\
        $\Delta$\,\%  & \textcolor{red}{$-$46\%} & \textcolor{red}{$+$50\%} & \textcolor{red}{$-$50\%} & \textcolor{red}{$+$77\%}  & \textcolor{red}{$-$58\%} & \textcolor{red}{$+$97\%} \\
        \bottomrule
    \end{tabular}%
    }
    \vspace*{-0.1in}
\end{table}

\paragraph{\textbf{Why smaller FLOPs can be slower and how to speed up GKA.}}
Although GKA reduces parameters and analytical FLOPs (Table~\ref{tab:param_efficiency}), in our current implementation, it is consistently slower and more memory-intensive (Table~\ref{tab:runtime_efficiency}). The key reason is kernel fusion: dot-product attention can use fused SDPA/FlashAttention kernels~\cite{dao2022flashattention} that avoid materializing the full $N\times N$ attention matrix, streaming tiles through on-chip memory. GKA requires forming pairwise squared distances and Gaussian affinities followed by separate exponentiation and row normalization, which (in our implementation) materializes $N\times N$ intermediates and increases memory traffic. However, it is possible to significantly speed up Gaussian kernel computation using the fast Gauss transform \cite{yang2004}, Nystrom approximation \cite{drineas2005nystrom,li2014large}, or GPU-based acceleration \cite{srinivasan2010gpuml}.

\paragraph{Scaling trends.}
Runtime penalties grow with model scale and batch size. From Tiny to Base, the inference throughput gap increases from $-$12\% to $-$46\% and GKA-B’s optimal inference batch size drops to 32 (vs.\ 64 for DeiT-B), consistent with larger activation footprints. In training, DeiT-S nearly doubles throughput when increasing per-GPU batch size from 64 to 128 (10{,}010 $\rightarrow$ 20{,}100 img/s), while GKA-S improves marginally (6{,}519 $\rightarrow$ 7{,}394 img/s), consistent with a memory-bandwidth bottleneck. At Base scale, GKA reaches 19.9\,GB per GPU at batch size 128, limiting further scaling without accumulation/checkpointing.

\paragraph{Discussion and outlook.}
These results highlight a gap between \emph{parametric} efficiency and \emph{wall-clock} efficiency: removing three projection GEMMs trades tensor-core-friendly matrix multiplications for bandwidth-dominated distance, exponentiation or normalization. A promising direction is fused Gaussian attention kernels (CUDA/Triton) that tile distance computation, exponentiation, masking, and normalization in a single IO-aware pass, analogous in spirit to FlashAttention~\cite{dao2022flashattention}. This systems optimization is orthogonal to the modeling contribution of GKA.

\subsection{Ablation Studies}
\label{sec:ablations}

\paragraph{Partial projection removal.}
To isolate which projections matter, we compare against Value-Less Transformers (VLT), which remove $W_V$ but retain $W_Q,W_K$ with dot-product attention. GKA removes all three input projections. Table~\ref{tab:ablation_projections} reports results for these models.

\begin{table}[t]
    \centering
    \caption{\textbf{Ablation of projection matrices} in ViT-Ti on ImageNet-1K. Standard: all projections. VLT: removes $W_V$. GKA: removes $W_Q,W_K,W_V$ and uses Gaussian similarity. All models are trained for 300 epochs with identical settings.}
    \label{tab:ablation_projections}
    \vspace{2mm}
    \begin{tabular}{l|c|c|c|c}
        \toprule
        \textbf{Variant} & \textbf{Projections} & \textbf{Params (M)} & \textbf{FLOPs (G)} & \textbf{Top-1 (\%)} \\
        \midrule
        DeiT-Ti (Standard) & $W_Q, W_K, W_V, W_O$ & 5.72 & 2.51 & 72.2 \\
        \textbf{VLT-Ti }            & $W_Q, W_K, W_O$       & 5.28 & 2.33 & \textbf{73.5} \\
        \textbf{GKA-Ti}             & $W_O$ only             & \textbf{4.38} & \textbf{1.98} & 71.54 \\
        \bottomrule
    \end{tabular}
    \vspace*{-0.1in}
\end{table}

VLT-Ti improves over DeiT-Ti by 1.3\,pp (73.5 vs.\ 72.2), suggesting that $W_V$ can be unnecessary or even harmful at this scale. Removing $W_Q$ and $W_K$ as well yields a modest accuracy drop (GKA-Ti at 71.54\%) while providing the largest reductions in parameters and FLOPs. Taken together, these results indicate that (i) the value projection is the easiest to remove, and (ii) learned query/key projections provide a measurable but non-essential benefit via learned similarity geometry that a fixed Gaussian kernel only partially recovers at this scale.

\section{Experiments with the NLP Task}
\label{sec:nlp}

\paragraph{Evaluation setting: \texttt{nanochat}.}
All language experiments use \texttt{nanochat}~\cite{nanochat_repo}, which provides end-to-end training and evaluation for GPT-style models. nanochat reports language-model quality in bits-per-byte (BPB; lower is better) and downstream capability using centered per-task scores and an aggregate CORE score aligned with DataComp-LM (DCLM)~\cite{dclm_paper,dclm_repo}. We follow nanochat's reference pipeline based on FineWeb-Edu~\cite{finewebedu} for tokenization and pretraining. All runs use vocabulary size 32{,}768 and sequence length 2{,}048.

\subsection{Models}

\paragraph{Baseline: standard causal self-attention.}
The baseline is the default depth-20 nanochat GPT-style Transformer~\cite{nanochat_repo} with learned $W_Q,W_K,W_V$ and output projection $W_O$.

\paragraph{Gaussian Kernel Attention (GKA, ours).}
We change only the attention operator: dot-product attention is replaced by a Gaussian RBF kernel operating directly on per-head residual-stream features, while keeping the surrounding block structure (norms, residuals, MLP) unchanged. Concretely, we (i) remove $W_Q,W_K,W_V$, (ii) apply RoPE and per-head normalization consistent with nanochat's QK normalization~\cite{nanochat_repo}, and (iii) compute masked, row-normalized Gaussian affinities:
% \begin{equation}
% K^{(h)}_{ij}=\exp\!\left(-\frac{\|x_i^{(h)}-x_j^{(h)}\|_2^2}{2\sigma_h^2}\right),\qquad
% W^{(h)}_{ij}=\frac{K^{(h)}_{ij}\,\mathbb{I}[\mathrm{allowed}(i,j)]}{\sum_{j'}K^{(h)}_{ij'}\,\mathbb{I}[\mathrm{allowed}(i,j')]},
% \end{equation}
\begin{equation}
K^{(h)}_{ij}=\exp\!\left(-\frac{\|\tilde{x}_i^{(h)}-\tilde{x}_j^{(h)}\|_2^2}{2\sigma_h^2}\right),\qquad
W^{(h)}_{ij}=\frac{K^{(h)}_{ij}\,\mathbb{I}[\mathrm{allowed}(i,j)]}{\sum_{j'}K^{(h)}_{ij'}\,\mathbb{I}[\mathrm{allowed}(i,j')]},
\end{equation}
where $\sigma_h=\exp(\log\sigma_h)$ is a learnable per-head bandwidth. The head output is $y_i^{(h)}=\sum_j W^{(h)}_{ij}\,\tilde{x}_j^{(h)}$, followed by the standard output projection.

\subsection{Depth-20 configurations}

We compare two depth-20 configurations using the same sequence length (2{,}048) and the same sliding-window pattern \texttt{SSSL}:
\begin{itemize}
  \item \textbf{Baseline (\texttt{nanochat\_base}, depth 20):} standard attention as implemented in nanochat~\cite{nanochat_repo}.
  \item \textbf{GKA (depth 20):} $n_{\text{layer}}=20$, $n_{\text{head}}=10$, $d_{\text{model}}=1280$, vocab size 32{,}768, window pattern \texttt{SSSL}.
\end{itemize}

\subsection{Results on the NLP Task}

\subsubsection{Language modeling quality (BPB)}

\begin{table}[htbp!]
\vspace*{-.3in}
\centering
\caption{\textbf{Depth-20 language modeling results (BPB; lower is better).}}
\label{tab:bpb}
\begin{tabular}{c|c|c|c}
\hline
Model & Train BPB & Val BPB & (Val--Train) \\
\hline
\texttt{nanochat\_base} (depth 20) & 0.7800 & 0.7884 & 0.0084 \\
\textbf{GKA (depth 20)} & 0.818256 & 0.819080 & 0.000824 \\
\hline
\end{tabular}
\vspace*{-0.2in}
\end{table}

From Table \ref{tab:bpb}, we observe that at depth 20, GKA yields higher BPB than the baseline. Notably, the train--validation gap is much smaller for GKA at this training point, consistent with reduced overfitting and/or under-training relative to the baseline.

\subsubsection{Downstream evaluation: CORE subset and task breakdown}

Nanochat reports centered per-task scores and a CORE aggregate over a fixed suite~\cite{nanochat_repo,dclm_paper}. For the GKA run, long-context tasks were not executed; therefore its reported ``CORE'' is computed over a reduced subset and is not directly comparable to the baseline CORE. We instead compare the shared subset of tasks and report the mean-centered score over this shared subset in Table \ref{tab:core_subset}.

\begin{table}[htbp!]
\vspace*{-0.2in}
\centering
\caption{\textbf{Centered CORE subset at depth 20.} GKA ``CORE'' is computed over this subset because long-context tasks were not run.}
\label{tab:core_subset}
\begin{tabular}{l|c|c|c}
\hline
Task & Baseline (centered) & \textbf{GKA (centered)} & $\Delta$ (GKA--Base) \\
\hline
hellaswag\_zeroshot \cite{hellaswag} & 0.2650 & 0.2267 & -0.0383 \\
copa \cite{copa}                    & 0.3000 & 0.3000 & \phantom{-}0 \\
openbook\_qa \cite{openbookqa}      & 0.1520 & \textbf{0.2000} & \phantom{-}0.0480 \\
lambada\_openai \cite{lambada}      & 0.3627 & 0.2750 & -0.0877 \\
winograd \cite{winograd}            & 0.3040 & 0.2800 & -0.0240 \\
winogrande \cite{winogrande}        & 0.0671 & \textbf{0.0900} & \phantom{-}0.0229 \\
\hline
Mean over shared subset             & 0.2418 & 0.2286 & -0.0132 \\
\hline
\end{tabular}
\vspace*{-0.2in}
\end{table}

On the shared subset, GKA improves OpenBookQA and Winogrande, matches COPA, and lags on LAMBADA and HellaSwag, yielding a modestly lower mean centered score.

\subsubsection{Model size and training compute}

\begin{table}[htbp!]
% \vspace*{-.4in}
\centering
\caption{\textbf{Depth-20 size and compute summary.}}
\label{tab:compute}
\begin{tabular}{l|c|c|c|c}
\hline
Model & Params (in M) & FLOPs/token & Train tokens & Total train FLOPs \\
\hline
\texttt{nanochat\_base} & 896 & 3.0042e9 & 4.5686e9 & 1.3725e19 \\
\textbf{GKA} & \textbf{378} & \textbf{2.4143e9} & \textbf{2.7787e9} & \textbf{6.7088e18} \\
\hline
% \vspace*{-0.3in}
\end{tabular}
\end{table}

Relative to the baseline, we observe from Table \ref{tab:compute} that the GKA depth-20 run uses $0.42\times$ parameters and $0.49\times$ total training FLOPs. Accordingly, the observed BPB and downstream differences reflect both the attention mechanism change and a substantial scale/compute reduction, motivating controlled scaling comparisons in future work.

\section{Summary and Discussion}
\label{sec:conclusion}

We introduced \emph{Gaussian Kernel Attention} (GKA), a projection-free attention mechanism that replaces the learned $W_Q$, $W_K$, and $W_V$ matrices with Gaussian RBF affinities parameterized by a single learnable bandwidth per head. GKA remains a drop-in replacement for multi-head attention, preserving the Transformer block interface and the output projection $W_O$.

Across vision and language, much of the performance of projection-based attention can be recovered with this simplified diffusion operator, at modest accuracy cost. In vision, GKA-Ti and GKA-S close to within 0.66 and 1.57 top-1 points of their DeiT counterparts on ImageNet-1K while reducing parameters by 23--25\% and FLOPs by 21--24\%. Projection ablations reveal a non-monotonic effect: removing $W_V$ can improve accuracy at small scale, while removing $W_Q$ and $W_K$ incurs only limited degradation for substantially larger efficiency gains.

In language modeling, a depth-20 GKA configuration trained at $0.42\times$ parameters and $0.49\times$ total training FLOPs relative to the baseline trains stably with a near-zero train--validation gap at the measured point. On the shared CORE subset, GKA improves OpenBookQA~\cite{openbookqa} and Winogrande~\cite{winogrande}, matches COPA, and trails on LAMBADA~\cite{lambada} and HellaSwag~\cite{hellaswag}. Although these comparisons are not compute-matched, the pattern is consistent with projection-free diffusion preserving relational aggregation while offering less flexible learned similarity geometry.

Conceptually, GKA makes explicit that self-attention is normalized kernel regression over tokens, linking Transformers to non-local filtering and kernel smoothing operators~\cite{buades2005nonlocal,wang2018nonlocal}. The learned bandwidths $\sigma_h$ expose an interpretable, continuous locality scale per head—an explicit inductive bias not directly parameterized in dot-product attention.

\paragraph{Limitations and future directions.}
GKA remains quadratic in sequence length (when unmasked) and cannot leverage fused SDPA/FlashAttention kernels~\cite{dao2022flashattention}, which explains the wall-clock gap despite fewer parameters and FLOPs. A clear systems direction is fused Gaussian attention kernels that tile distance computation, exponentiation, masking, and normalization in a single IO-aware pass. On the modeling side, compute-matched scaling studies are needed to isolate architectural effects, and hybrid designs that interleave GKA with projected attention may recover additional expressivity while retaining most parameter and interpretability benefits.

Overall, GKA demonstrates that projection-based similarity is not a prerequisite for effective token interaction. A minimal Gaussian diffusion operator, with one learned scalar per head, recovers most of the accuracy of standard attention while substantially reducing parameters and FLOPs. By isolating the role of learned projection geometry, GKA sharpens our understanding of what attention actually contributes and establishes a new, interpretable point in the Transformer accuracy--efficiency design space.

% In your main paper file:
% \bibliographystyle{plainnat}
% \bibliography{references}

% \clearpage\mbox{}Page \thepage\ of the manuscript.
% \clearpage\mbox{}Page \thepage\ of the manuscript.
% \clearpage\mbox{}Page \thepage\ of the manuscript.
% \clearpage\mbox{}Page \thepage\ of the manuscript.
% \clearpage\mbox{}Page \thepage\ of the manuscript. This is the last page.
% Now we have reached the maximum length of an ECCV \ECCVyear{} submission (excluding references and acknowledgements).
% References should start immediately after the main text, but can continue past p.\ 14 if needed. 
% \clearpage  % TODO FINAL: This \clearpage needs to be removed from both review and camera-ready versions.

% \section*{Acknowledgements}
% Please insert your acknowledgments here.

\bibliographystyle{splncs04}
\bibliography{reference}

\section{Attention Visualization Methodology}
\label{sec:supp_attention_vis}

To provide interpretable insight into the behavior of Gaussian Kernel Attention (GKA), we construct an exhaustive visualization pipeline that probes attention patterns at multiple levels of granularity: per-head, per-layer, patch-to-patch, and network-wide. All visualizations are generated from a single forward pass through the trained GKA model. In this documents we show the results for GKA-Tiny model for a single image. For other models and images please refer the supplementary folder.

\subsection{Gaussian Kernel Attention Recap}
\label{sec:supp_gka_recap}

Recall that GKA replaces the standard dot-product attention mechanism with a Gaussian radial basis function (RBF) kernel operating directly on token features. Given input tokens $\mathbf{X} \in \mathbb{R}^{N \times d}$ reshaped into $H$ heads of dimension $d_h = d/H$, the attention weights for head $h$ are computed as:
\begin{equation}
    A_{ij}^{(h)} = \frac{\exp\!\bigl(-\|\mathbf{x}_i^{(h)} - \mathbf{x}_j^{(h)}\|^2 \,/\, 2{\sigma^{(h)}}^2\bigr)}{\sum_{k=1}^{N} \exp\!\bigl(-\|\mathbf{x}_i^{(h)} - \mathbf{x}_k^{(h)}\|^2 \,/\, 2{\sigma^{(h)}}^2\bigr)},
    \label{eq:gka_attn}
\end{equation}
where $\sigma^{(h)}$ is a learnable per-head bandwidth parameter, and the pairwise squared Euclidean distances are computed as $\|\mathbf{x}_i - \mathbf{x}_j\|^2 = \|\mathbf{x}_i\|^2 + \|\mathbf{x}_j\|^2 - 2\,\mathbf{x}_i^\top \mathbf{x}_j$. Unlike standard multi-head self-attention~\cite{vaswani2017attention}, GKA eliminates all learned $W_Q$, $W_K$, and $W_V$ projection matrices, retaining only the per-head scalar $\sigma^{(h)}$ and the output projection $W_O$.

\subsection{Attention Rollout}
\label{sec:supp_rollout}

Following~\cite{abnar2020quantifying}, we compute attention rollout to approximate the total attention flow from the input to the \texttt{[CLS]} token across all layers. At each layer $\ell$, we first compute the head-averaged attention matrix $\bar{A}^{(\ell)} = \frac{1}{H}\sum_{h} A^{(\ell,h)}$, then augment it with a residual connection identity term:
\begin{equation}
    \hat{A}^{(\ell)} = 0.5 \cdot \bar{A}^{(\ell)} + 0.5 \cdot I,
    \label{eq:rollout_residual}
\end{equation}
followed by row-normalization. The cumulative rollout is then $R = \prod_{\ell=1}^{L} \hat{A}^{(\ell)}$, and the CLS row $R_{0, 1:N}$ is reshaped and overlaid on the input image. To examine the effect of pruning low-confidence connections, we additionally apply discard-ratio thresholding~\cite{abnar2020quantifying} at ratios $\{0.0, 0.5, 0.9\}$: entries below the corresponding quantile are zeroed before re-normalization. This reveals whether the network concentrates attention on semantically meaningful regions when weak connections are removed.

\begin{figure*}[tbh]
    \centering
    \includegraphics[width=01\linewidth]{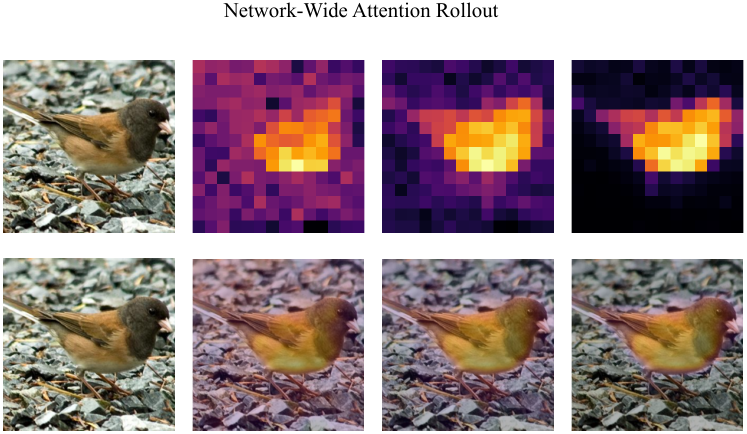}
\caption{}
\label{fig:vlt-attention}
\end{figure*}
%\clearpage
\subsection{Learned Bandwidth ($\sigma$) Analysis}
\label{sec:supp_sigma}

The per-head bandwidth parameters $\{\sigma^{(\ell, h)}\}$ are the \emph{only} learnable parameters within GKA's attention computation (aside from $W_O$). We visualize these in two complementary formats:

\paragraph{Heatmap.} A $H \times L$ matrix displaying the learned $\sigma$ value for each head--layer pair, colored using the \texttt{YlOrRd} (yellow--orange--red) sequential colormap. This reveals whether the network has learned to differentiate bandwidth across heads and layers.

\paragraph{Line plot.} Per-head $\sigma$ trajectories plotted against layer depth, showing the evolution of bandwidth from early to late layers. In standard dot-product attention, analogous analysis would track the effective temperature of the softmax; here, $\sigma$ directly controls the spatial extent of the Gaussian kernel, with larger values producing broader (more uniform) attention and smaller values producing sharper (more localized) attention.

\begin{figure*}[tbh]
    \centering
    \includegraphics[width=01\linewidth]{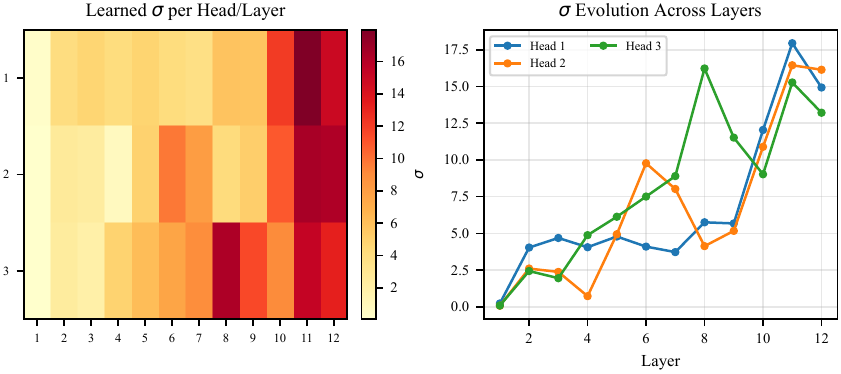}
\caption{}
\label{fig:vlt-attention}
\end{figure*}

\subsection{Per-Head CLS Attention Maps}
\label{sec:supp_per_head}

For each layer $\ell \in \{1, \ldots, L\}$ and head $h \in \{1, \ldots, H\}$, we extract the attention distribution from the \texttt{[CLS]} token to all spatial patch tokens. Given an input image of size $224 \times 224$ with patch size $16 \times 16$, this yields a $14 \times 14$ spatial attention map per head. Formally, for layer $\ell$ and head $h$, the CLS-to-patch attention is:
\begin{equation}
    \mathbf{M}_{\text{CLS}}^{(\ell, h)} = A^{(\ell, h)}_{0, 1:N} \in \mathbb{R}^{P \times P},
    \label{eq:cls_attn}
\end{equation}
where $P = 14$ is the spatial grid dimension and index $0$ corresponds to the \texttt{[CLS]} token. Each $14 \times 14$ map is bilinearly upscaled to the original image resolution and overlaid with partial transparency ($\alpha{=}0.6$) using the \texttt{inferno} colormap, following the visualization conventions of~\cite{dosovitskiy2021image,caron2021emerging}. We generate one composite figure per layer containing all $H$ heads side-by-side, yielding $L$ figures in total.

\begin{figure*}[tbh]
    \centering
    \includegraphics[width=01\linewidth]{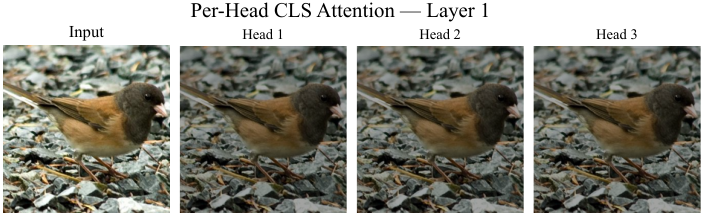}
\caption{}
\label{fig:vlt-attention}
\end{figure*}

\begin{figure*}[tbh]
    \centering
    \includegraphics[width=01\linewidth]{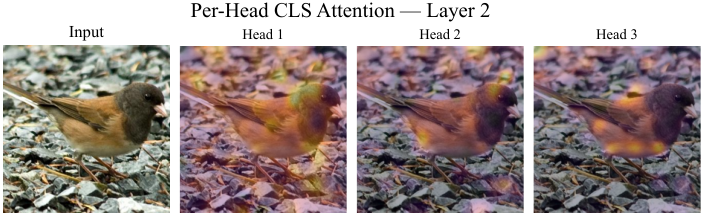}
\caption{}
\label{fig:vlt-attention}
\end{figure*}
\begin{figure*}[tbh]
    \centering
    \includegraphics[width=01\linewidth]{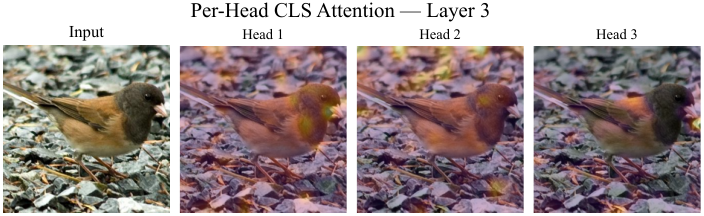}
\caption{}
\label{fig:vlt-attention}
\end{figure*}
\begin{figure*}[tbh]
    \centering
    \includegraphics[width=01\linewidth]{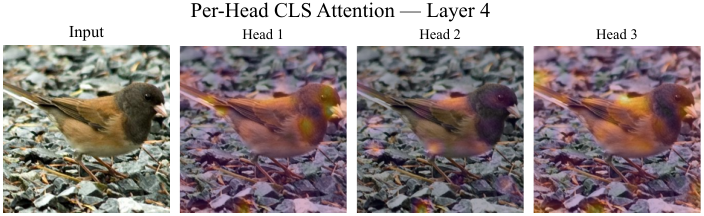}
\caption{}
\label{fig:vlt-attention}
\end{figure*}
\begin{figure*}[tbh]
    \centering
    \includegraphics[width=01\linewidth]{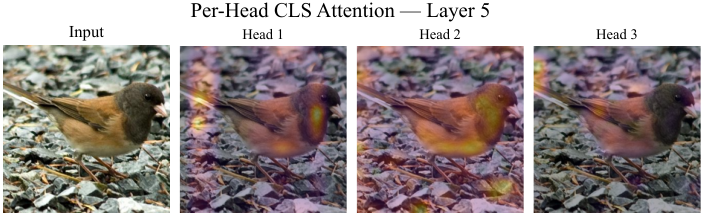}
\caption{}
\label{fig:vlt-attention}
\end{figure*}
\begin{figure*}[tbh]
    \centering
    \includegraphics[width=01\linewidth]{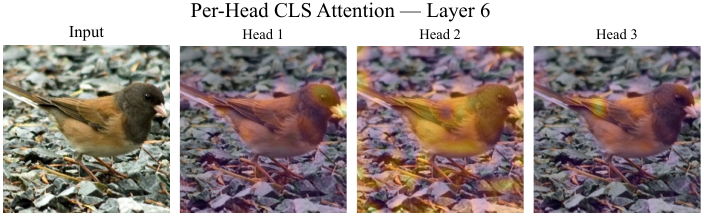}
\caption{}
\label{fig:vlt-attention}
\end{figure*}
\begin{figure*}[tbh]
    \centering
    \includegraphics[width=01\linewidth]{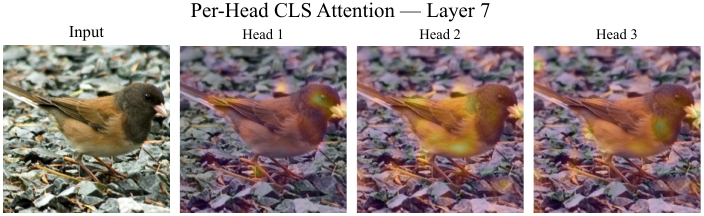}
\caption{}
\label{fig:vlt-attention}
\end{figure*}
\begin{figure*}[tbh]
    \centering
    \includegraphics[width=01\linewidth]{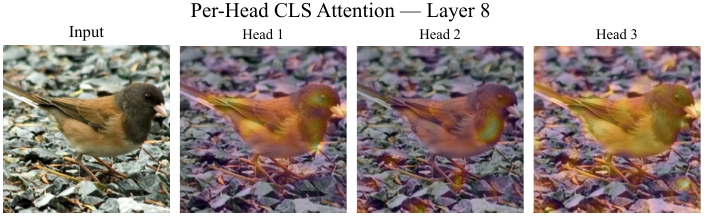}
\caption{}
\label{fig:vlt-attention}
\end{figure*}
\begin{figure*}[tbh]
    \centering
    \includegraphics[width=01\linewidth]{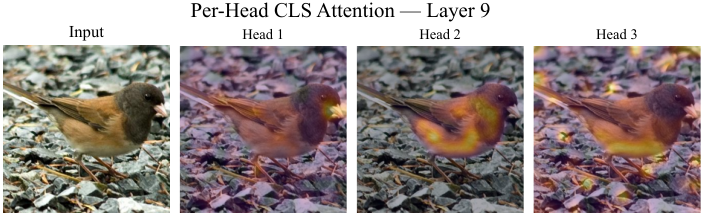}
\caption{}
\label{fig:vlt-attention}
\end{figure*}
\begin{figure*}[tbh]
    \centering
    \includegraphics[width=01\linewidth]{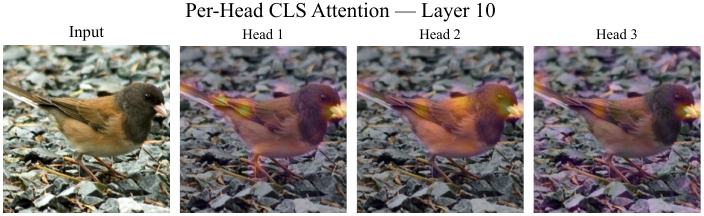}
\caption{}
\label{fig:vlt-attention}
\end{figure*}
\begin{figure*}[tbh]
    \centering
    \includegraphics[width=01\linewidth]{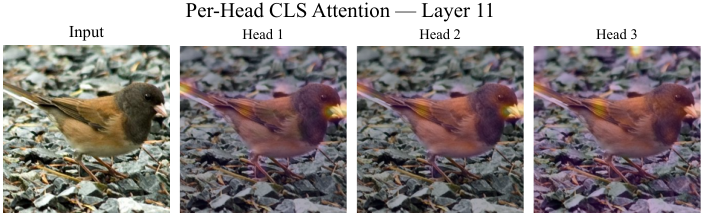}
\caption{}
\label{fig:vlt-attention}
\end{figure*}
\begin{figure*}[tbh]
    \centering
    \includegraphics[width=01\linewidth]{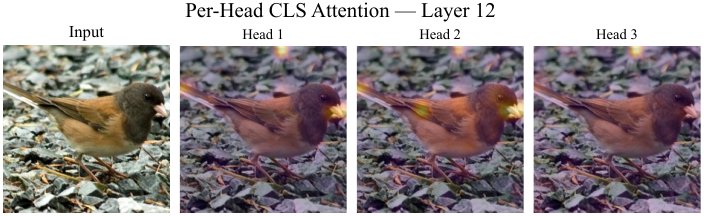}
\caption{}
\label{fig:vlt-attention}
\end{figure*}

\clearpage

\subsection{Patch-to-Patch Attention}
\label{sec:supp_patch_to_patch}

To examine how spatial tokens attend to one another (rather than how \texttt{[CLS]} aggregates information), we select four query patches at canonical positions---specifically at grid coordinates $(\lfloor P/4 \rfloor, \lfloor P/4 \rfloor)$, $(\lfloor P/2 \rfloor, \lfloor P/2 \rfloor)$, $(\lfloor 3P/4 \rfloor, \lfloor 3P/4 \rfloor)$, and $(\lfloor P/4 \rfloor, \lfloor 3P/4 \rfloor)$---corresponding to the upper-left, center, lower-right, and upper-right quadrants. For each query patch $q$, we compute the head-averaged attention distribution over all spatial tokens:
\begin{equation}
    \mathbf{M}_{\text{patch}}^{(\ell, q)} = \frac{1}{H} \sum_{h=1}^{H} A^{(\ell, h)}_{q, 1:N} \in \mathbb{R}^{P \times P}.
    \label{eq:patch_attn}
\end{equation}
Each per-query map is displayed both as a raw $14 \times 14$ heatmap and as an upscaled overlay on the original image, with the query patch location indicated by a colored bounding box. This visualization reveals the effective receptive field of individual patches and illustrates how GKA's distance-based kernel induces spatially localized or globally diffuse attention depending on the scene content.

\begin{figure*}[tbh]
    \centering
    \includegraphics[width=01\linewidth]{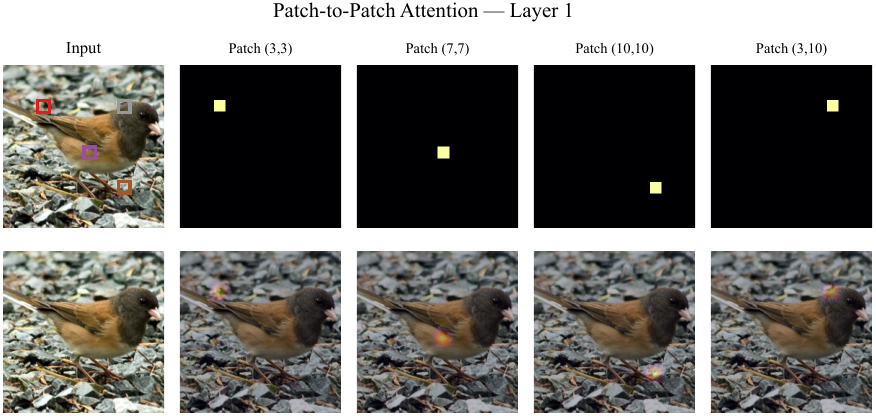}
\caption{}
\label{fig:vlt-attention}
\end{figure*}
\begin{figure*}[tbh]
    \centering
    \includegraphics[width=01\linewidth]{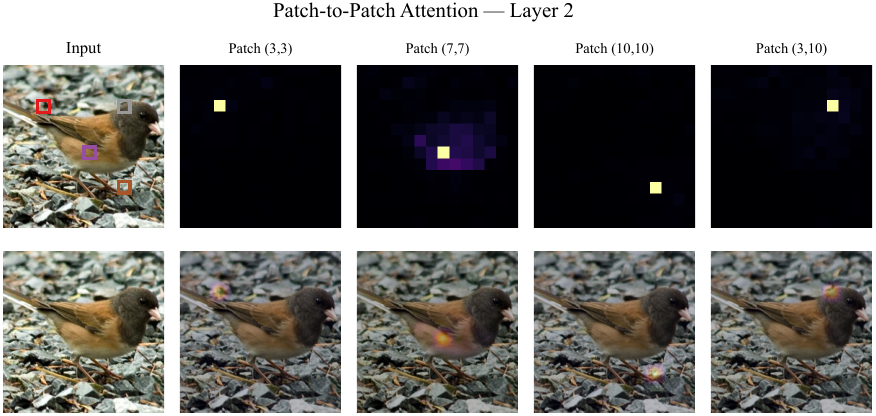}
\caption{}
\label{fig:vlt-attention}
\end{figure*}
\begin{figure*}[tbh]
    \centering
    \includegraphics[width=01\linewidth]{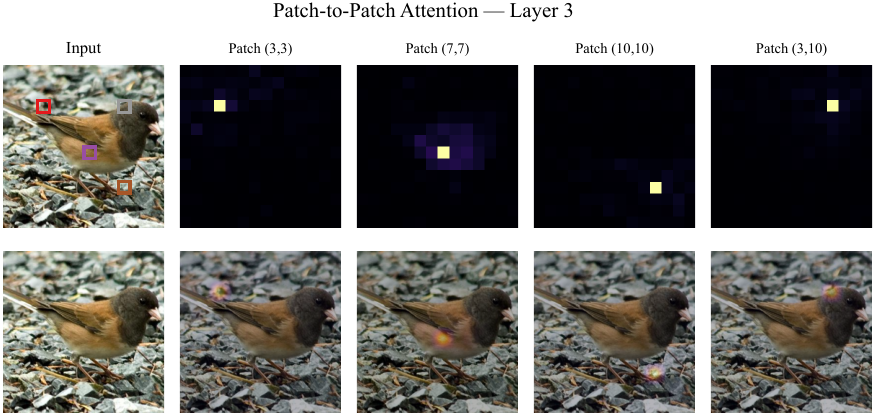}
\caption{}
\label{fig:vlt-attention}
\end{figure*}
\begin{figure*}[tbh]
    \centering
    \includegraphics[width=01\linewidth]{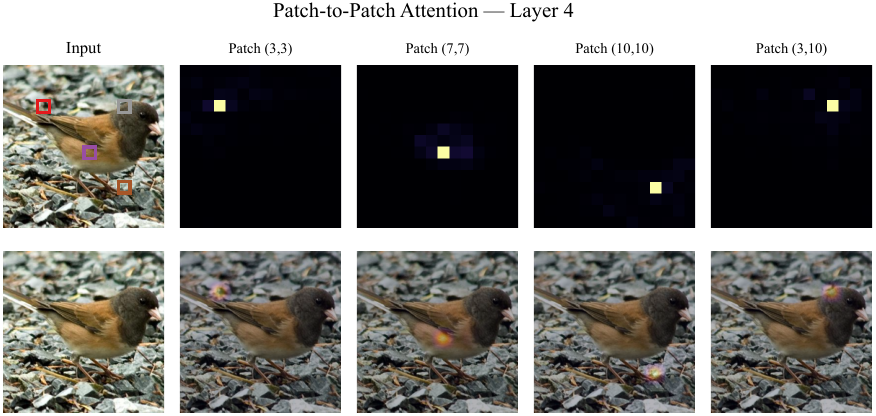}
\caption{}
\label{fig:vlt-attention}
\end{figure*}
\begin{figure*}[tbh]
    \centering
    \includegraphics[width=01\linewidth]{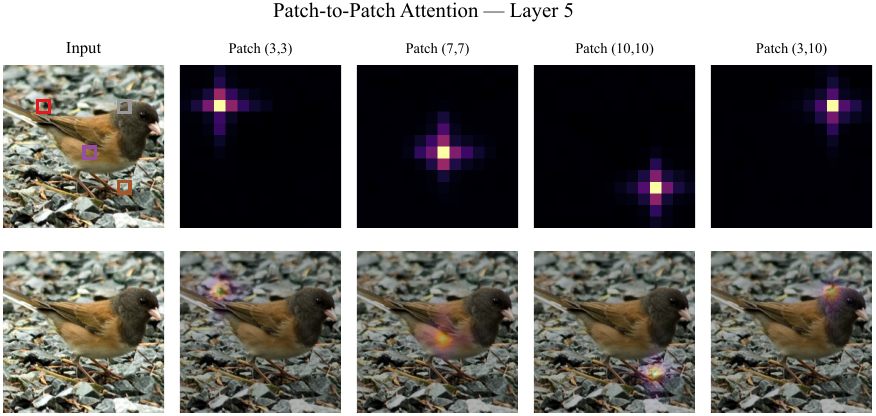}
\caption{}
\label{fig:vlt-attention}
\end{figure*}
\begin{figure*}[tbh]
    \centering
    \includegraphics[width=01\linewidth]{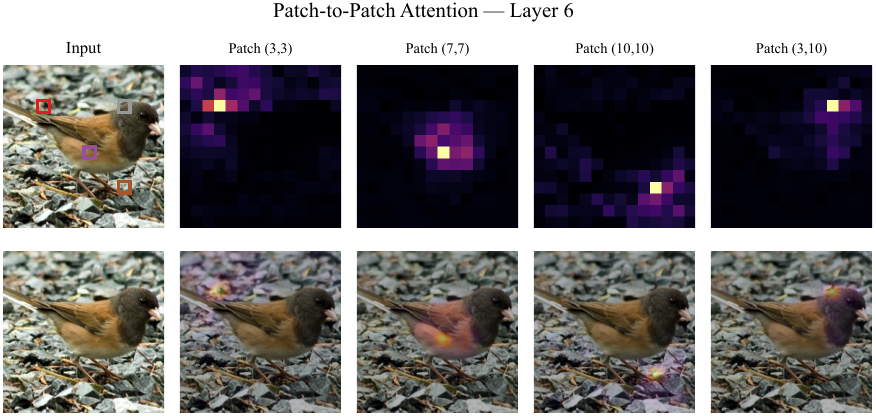}
\caption{}
\label{fig:vlt-attention}
\end{figure*}
\begin{figure*}[tbh]
    \centering
    \includegraphics[width=01\linewidth]{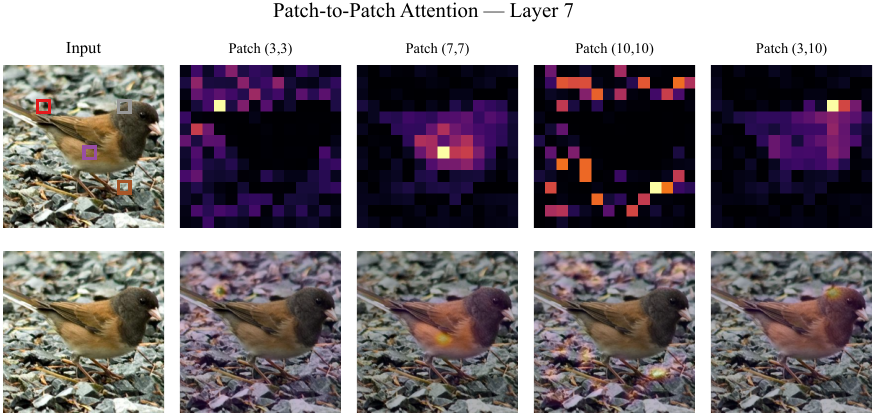}
\caption{}
\label{fig:vlt-attention}
\end{figure*}
\begin{figure*}[tbh]
    \centering
    \includegraphics[width=01\linewidth]{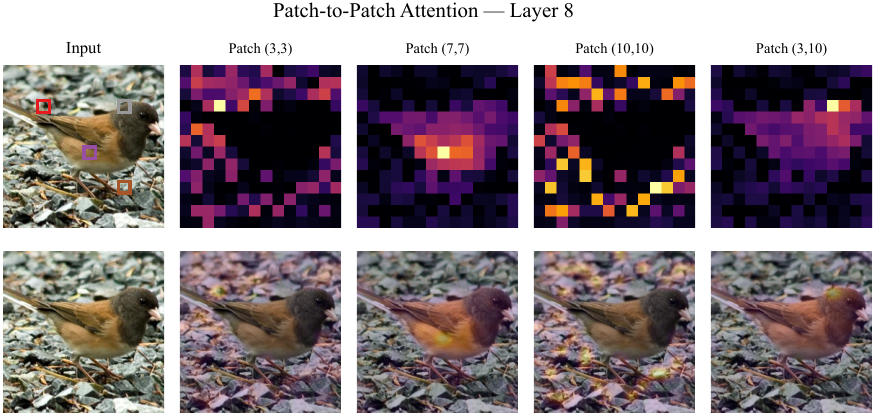}
\caption{}
\label{fig:vlt-attention}
\end{figure*}
\begin{figure*}[tbh]
    \centering
    \includegraphics[width=01\linewidth]{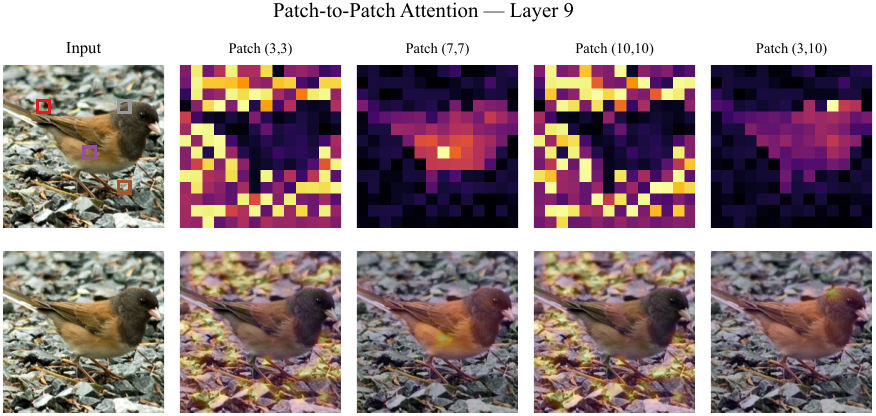}
\caption{}
\label{fig:vlt-attention}
\end{figure*}
\begin{figure*}[tbh]
    \centering
    \includegraphics[width=01\linewidth]{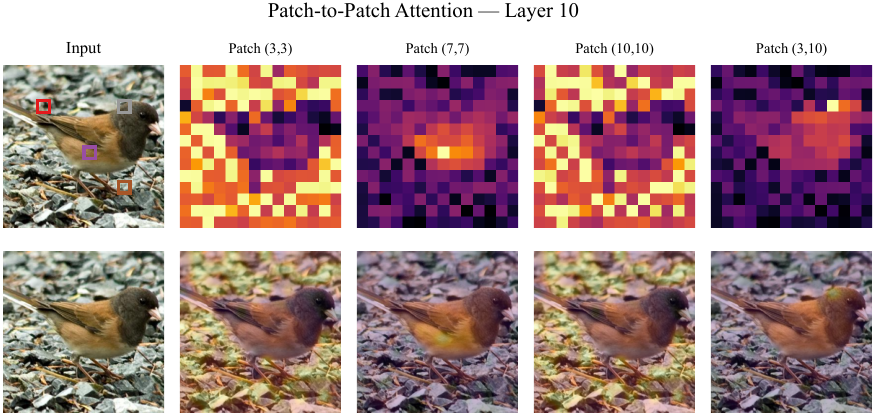}
\caption{}
\label{fig:vlt-attention}
\end{figure*}
\begin{figure*}[tbh]
    \centering
    \includegraphics[width=01\linewidth]{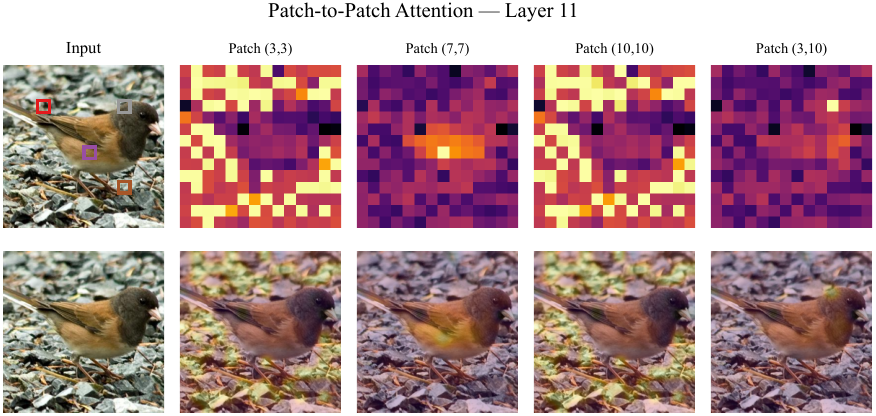}
\caption{}
\label{fig:vlt-attention}
\end{figure*}
\begin{figure*}[tbh]
    \centering
    \includegraphics[width=01\linewidth]{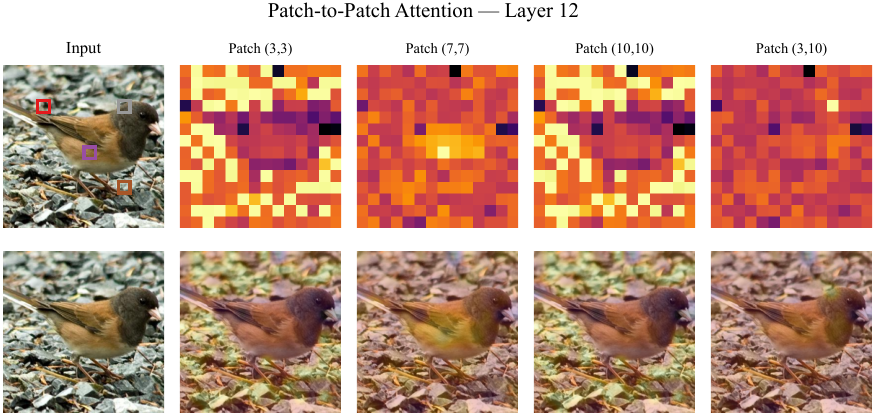}
\caption{}
\label{fig:vlt-attention}
\end{figure*}

\clearpage

\subsection{Layer-wise Kernel Evolution}
\label{sec:supp_kernel_evolution}

To provide a holistic view of how attention patterns develop across depth, we generate a single composite figure with $L$ rows (one per layer) and three columns: (1)~the head-averaged full attention matrix, (2)~the head-averaged CLS-to-patch attention as a $14 \times 14$ heatmap, and (3)~the same CLS attention overlaid on the input image. This three-column layout, spanning all $L{=}12$ layers, allows direct visual comparison of how the Gaussian kernel's effective receptive field evolves from early layers (which tend toward near-identity attention when $\sigma$ is small) to later layers.

\begin{figure*}[tbh]
    \centering
    \includegraphics[width=0.8\linewidth]{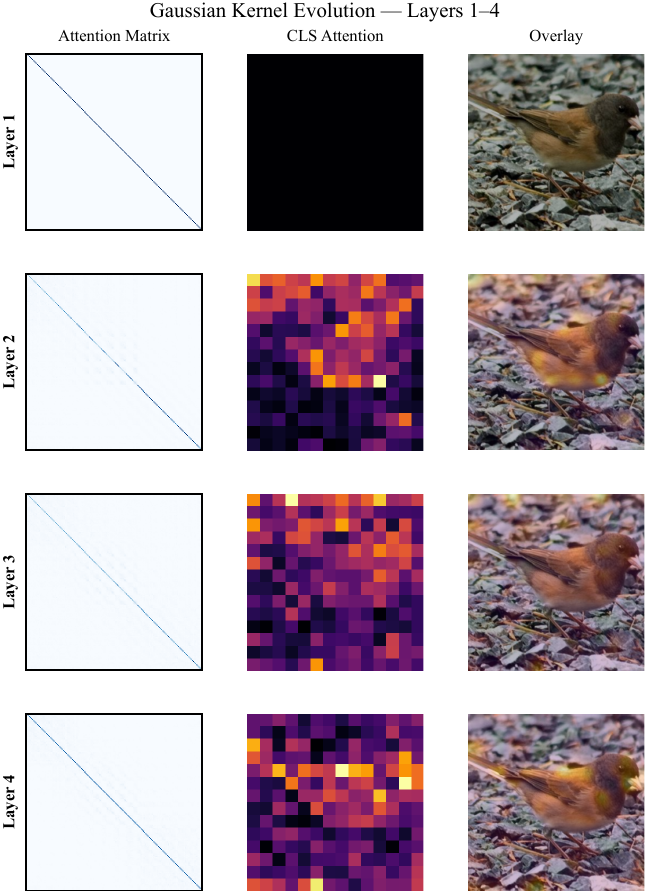}
\caption{}
\label{fig:vlt-attention}
\end{figure*}
\begin{figure*}[tbh]
    \centering
    \includegraphics[width=0.8\linewidth]{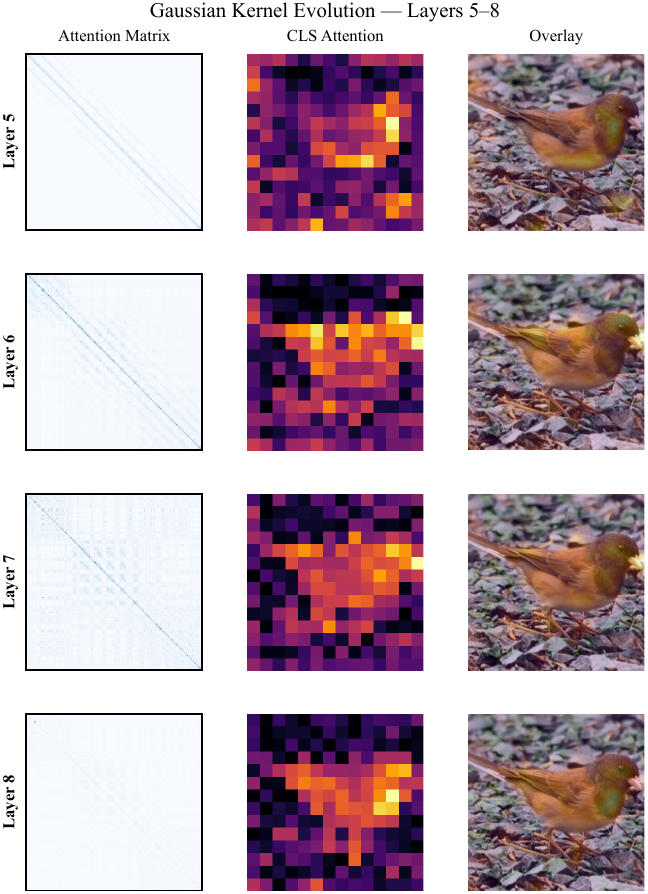}
\caption{}
\label{fig:vlt-attention}
\end{figure*}
\begin{figure*}[tbh]
    \centering
    \includegraphics[width=0.8\linewidth]{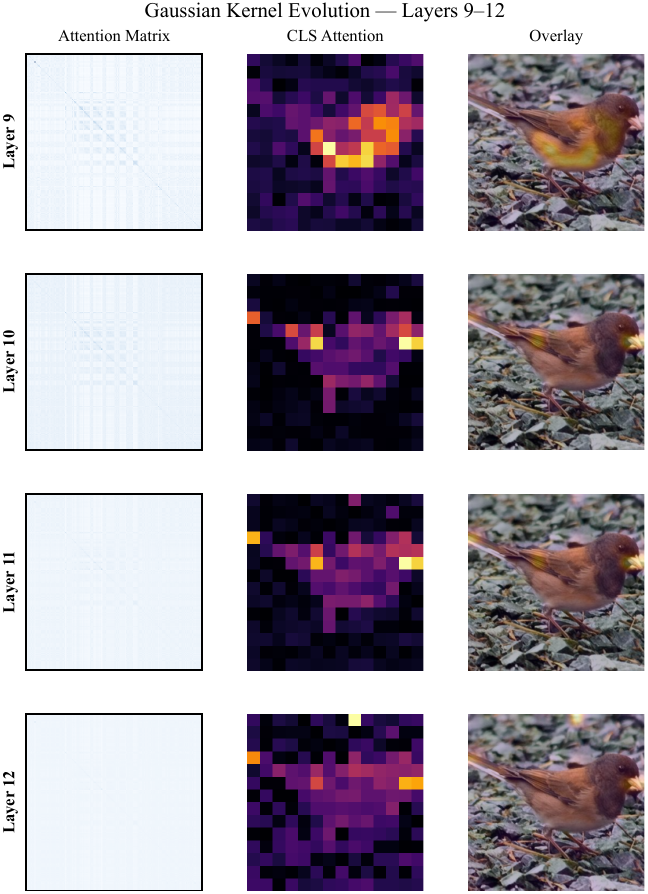}
\caption{}
\label{fig:vlt-attention}
\end{figure*}
\clearpage

\subsection{Raw Attention Matrices with Diagonal Masking}
\label{sec:supp_raw_matrices}

A distinctive property of GKA is that the self-attention weight $A_{ii}^{(h)}$ (the diagonal) tends to dominate when the learned bandwidth $\sigma^{(h)}$ remains close to its initialization value, since $\|\mathbf{x}_i - \mathbf{x}_i\|^2 = 0$ always yields the maximum kernel response. To reveal the structure of \emph{cross-token} attention, we visualize the full $N \times N$ attention matrix for each layer--head pair with the diagonal explicitly masked (set to zero) before applying the colormap. This rescales the color range to the off-diagonal entries, exposing fine-grained spatial relationships that would otherwise be invisible. For visualization clarity, matrices exceeding $50$ tokens are uniformly subsampled. We produce one figure per layer--head combination, yielding $L \times H$ individual matrices.

\begin{figure*}[tbh]
    \centering
    \includegraphics[width=01\linewidth]{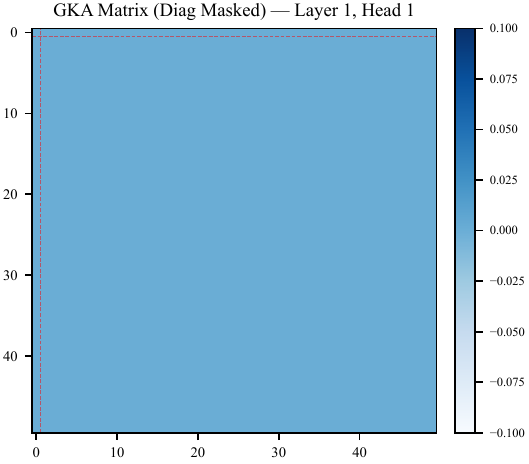}
\caption{}
\label{fig:vlt-attention}
\end{figure*}

\begin{figure*}[tbh]
    \centering
    \includegraphics[width=1\linewidth]{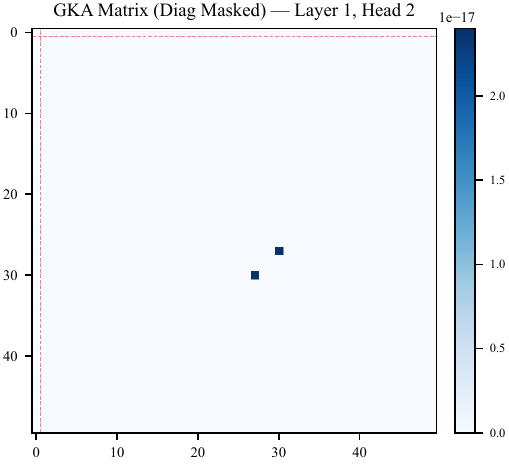}
\caption{}
\label{fig:vlt-attention}
\end{figure*}

\begin{figure*}[tbh]
    \centering
    \includegraphics[width=1\linewidth]{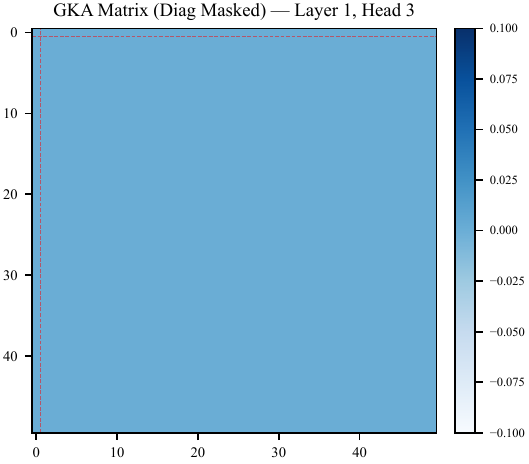}
\caption{}
\label{fig:vlt-attention}
\end{figure*}

\begin{figure*}[tbh]
    \centering
    \includegraphics[width=01\linewidth]{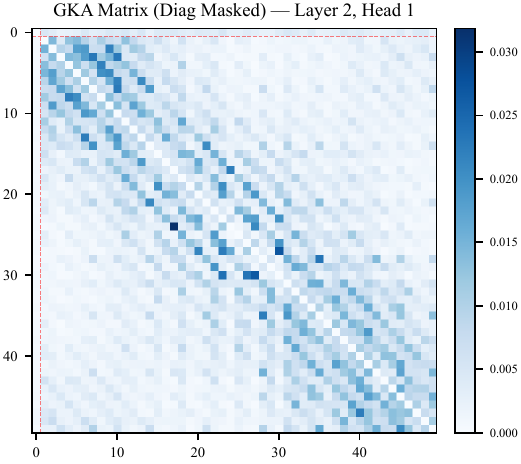}
\caption{}
\label{fig:vlt-attention}
\end{figure*}

\begin{figure*}[tbh]
    \centering
    \includegraphics[width=1\linewidth]{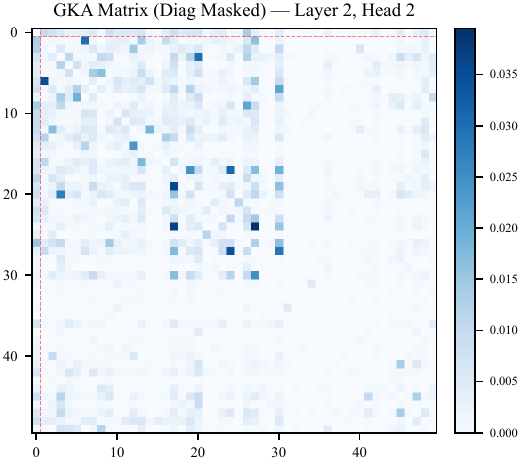}
\caption{}
\label{fig:vlt-attention}
\end{figure*}

\begin{figure*}[tbh]
    \centering
    \includegraphics[width=1\linewidth]{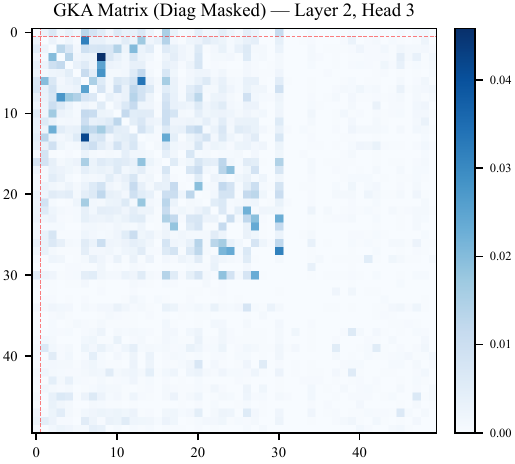}
\caption{}
\label{fig:vlt-attention}
\end{figure*}

\begin{figure*}[tbh]
    \centering
    \includegraphics[width=01\linewidth]{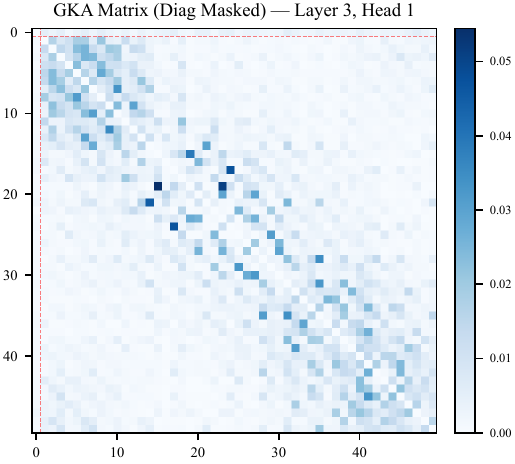}
\caption{}
\label{fig:vlt-attention}
\end{figure*}

\begin{figure*}[tbh]
    \centering
    \includegraphics[width=1\linewidth]{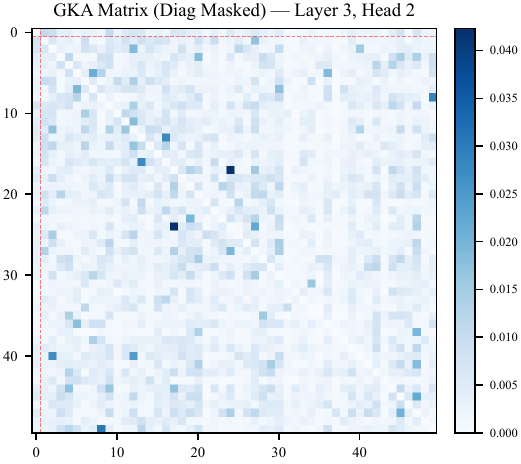}
\caption{}
\label{fig:vlt-attention}
\end{figure*}

\begin{figure*}[tbh]
    \centering
    \includegraphics[width=1\linewidth]{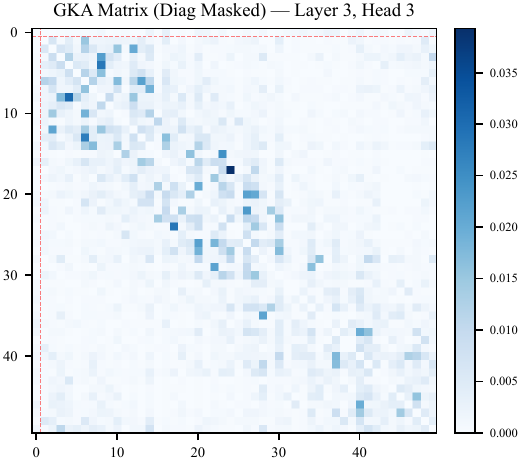}
\caption{}
\label{fig:vlt-attention}
\end{figure*}

\begin{figure*}[tbh]
    \centering
    \includegraphics[width=01\linewidth]{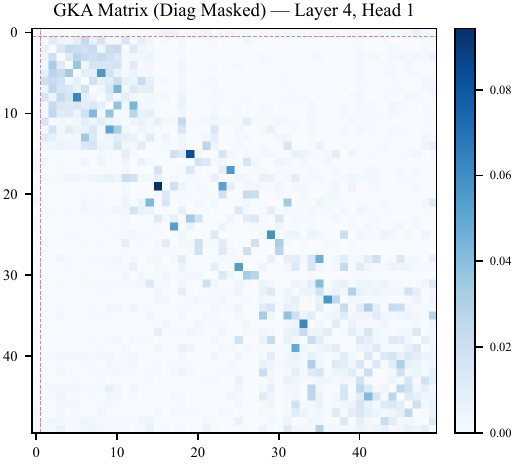}
\caption{}
\label{fig:vlt-attention}
\end{figure*}

\begin{figure*}[tbh]
    \centering
    \includegraphics[width=1\linewidth]{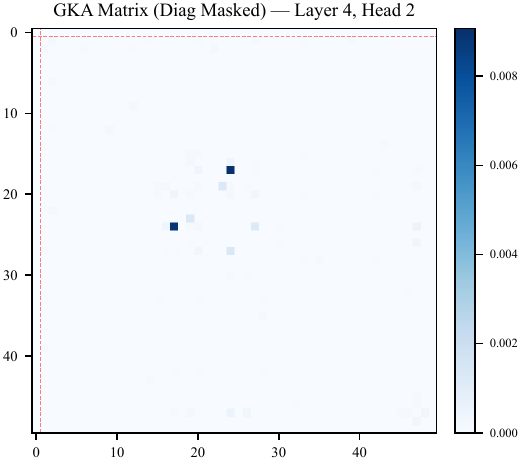}
\caption{}
\label{fig:vlt-attention}
\end{figure*}

\begin{figure*}[tbh]
    \centering
    \includegraphics[width=1\linewidth]{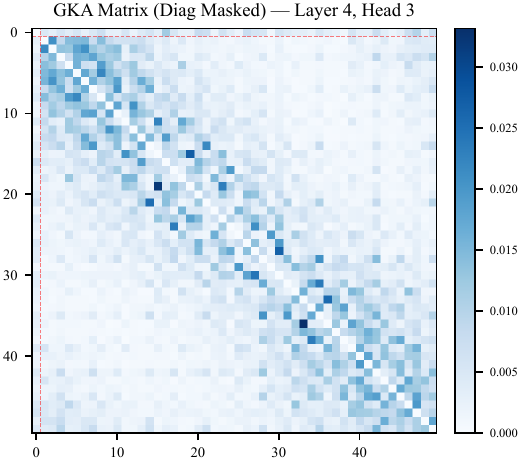}
\caption{}
\label{fig:vlt-attention}
\end{figure*}

\begin{figure*}[tbh]
    \centering
    \includegraphics[width=01\linewidth]{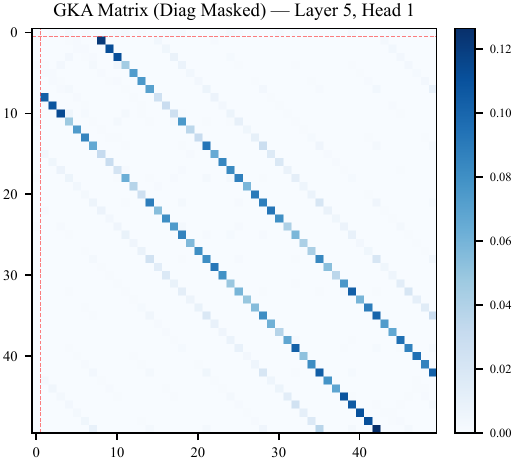}
\caption{}
\label{fig:vlt-attention}
\end{figure*}

\begin{figure*}[tbh]
    \centering
    \includegraphics[width=1\linewidth]{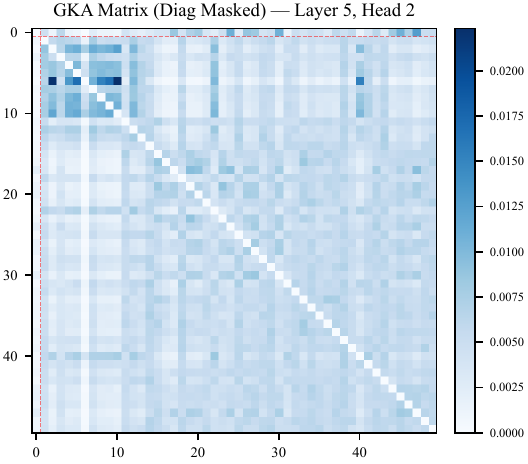}
\caption{}
\label{fig:vlt-attention}
\end{figure*}

\begin{figure*}[tbh]
    \centering
    \includegraphics[width=1\linewidth]{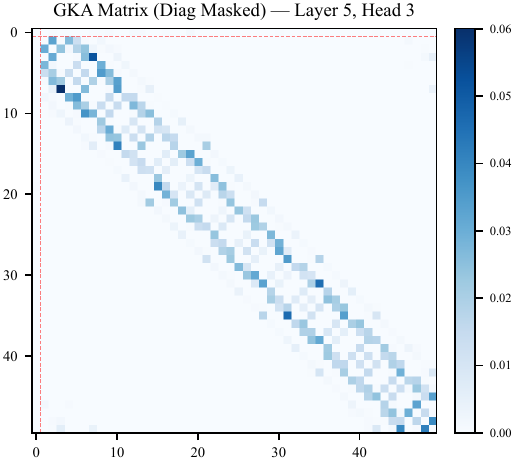}
\caption{}
\label{fig:vlt-attention}
\end{figure*}

\begin{figure*}[tbh]
    \centering
    \includegraphics[width=01\linewidth]{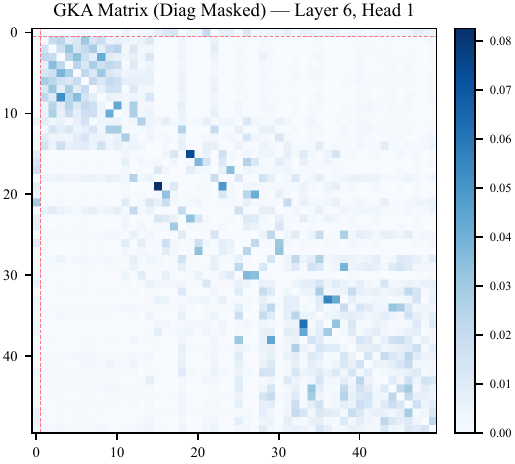}
\caption{}
\label{fig:vlt-attention}
\end{figure*}

\begin{figure*}[tbh]
    \centering
    \includegraphics[width=1\linewidth]{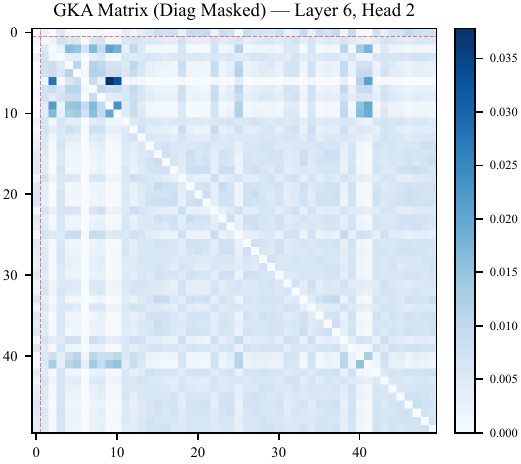}
\caption{}
\label{fig:vlt-attention}
\end{figure*}

\begin{figure*}[tbh]
    \centering
    \includegraphics[width=1\linewidth]{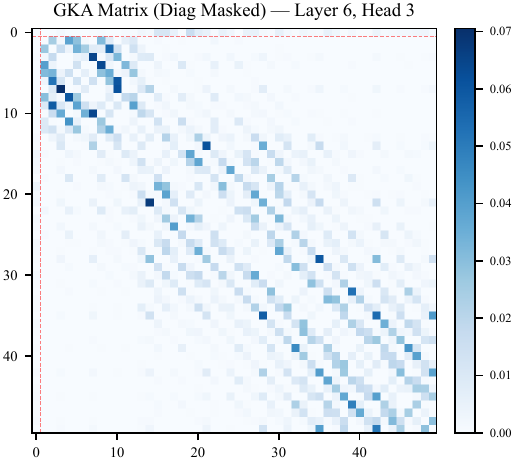}
\caption{}
\label{fig:vlt-attention}
\end{figure*}

\begin{figure*}[tbh]
    \centering
    \includegraphics[width=01\linewidth]{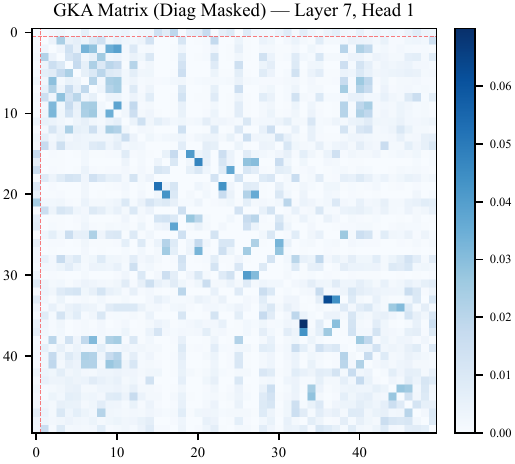}
\caption{}
\label{fig:vlt-attention}
\end{figure*}

\begin{figure*}[tbh]
    \centering
    \includegraphics[width=1\linewidth]{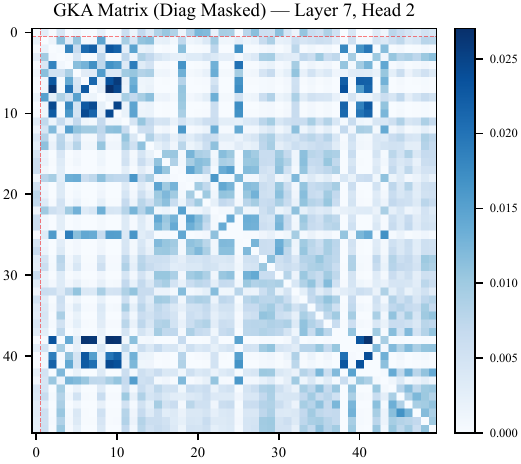}
\caption{}
\label{fig:vlt-attention}
\end{figure*}

\begin{figure*}[tbh]
    \centering
    \includegraphics[width=1\linewidth]{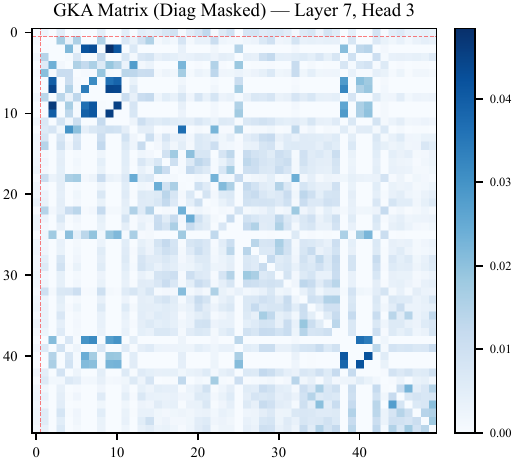}
\caption{}
\label{fig:vlt-attention}
\end{figure*}

\begin{figure*}[tbh]
    \centering
    \includegraphics[width=01\linewidth]{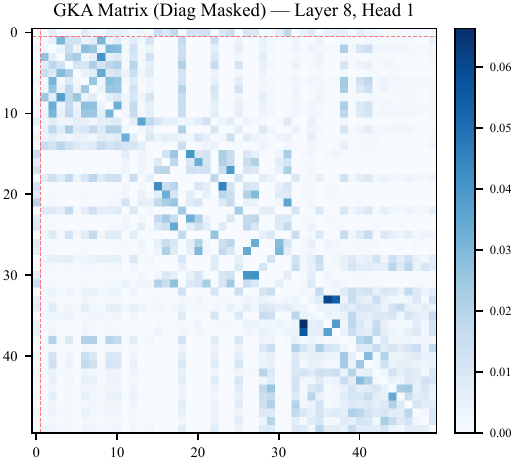}
\caption{}
\label{fig:vlt-attention}
\end{figure*}

\begin{figure*}[tbh]
    \centering
    \includegraphics[width=1\linewidth]{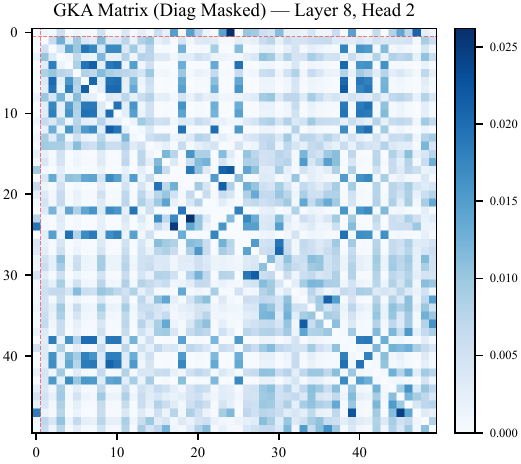}
\caption{}
\label{fig:vlt-attention}
\end{figure*}

\begin{figure*}[tbh]
    \centering
    \includegraphics[width=1\linewidth]{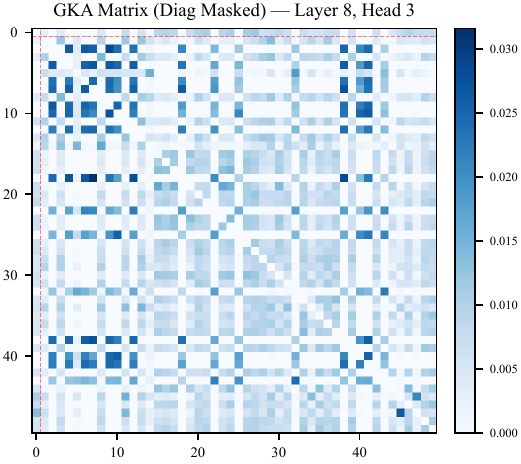}
\caption{}
\label{fig:vlt-attention}
\end{figure*}

\begin{figure*}[tbh]
    \centering
    \includegraphics[width=01\linewidth]{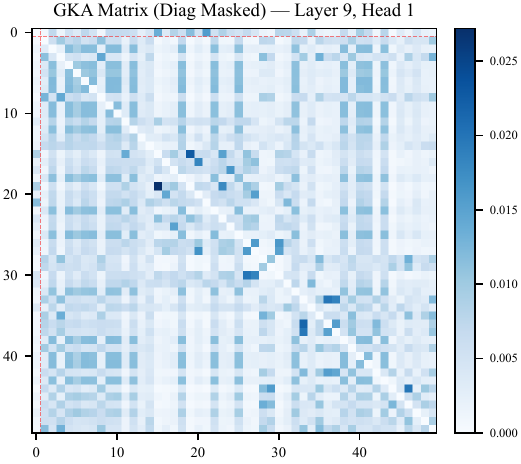}
\caption{}
\label{fig:vlt-attention}
\end{figure*}

\begin{figure*}[tbh]
    \centering
    \includegraphics[width=1\linewidth]{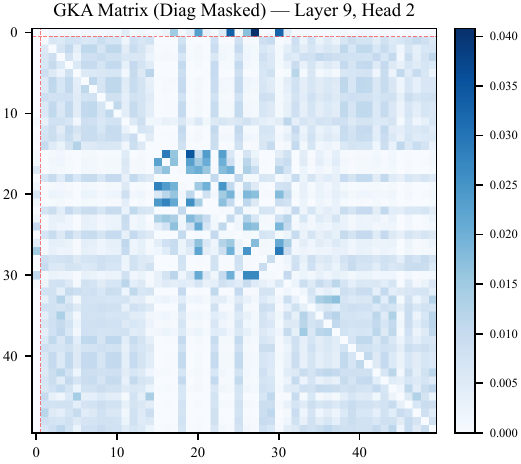}
\caption{}
\label{fig:vlt-attention}
\end{figure*}

\begin{figure*}[tbh]
    \centering
    \includegraphics[width=1\linewidth]{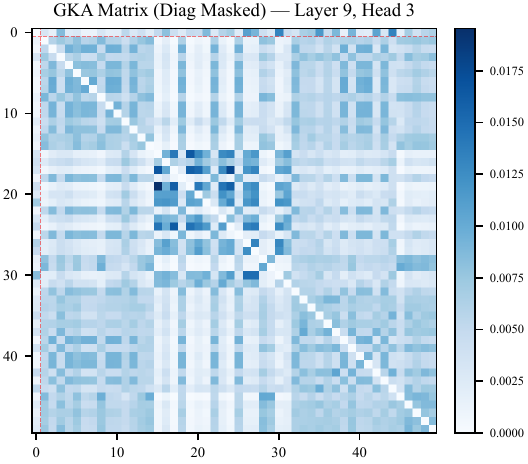}
\caption{}
\label{fig:vlt-attention}
\end{figure*}

\begin{figure*}[tbh]
    \centering
    \includegraphics[width=01\linewidth]{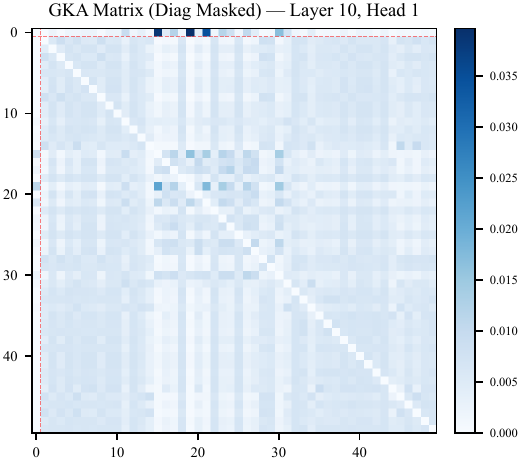}
\caption{}
\label{fig:vlt-attention}
\end{figure*}

\begin{figure*}[tbh]
    \centering
    \includegraphics[width=1\linewidth]{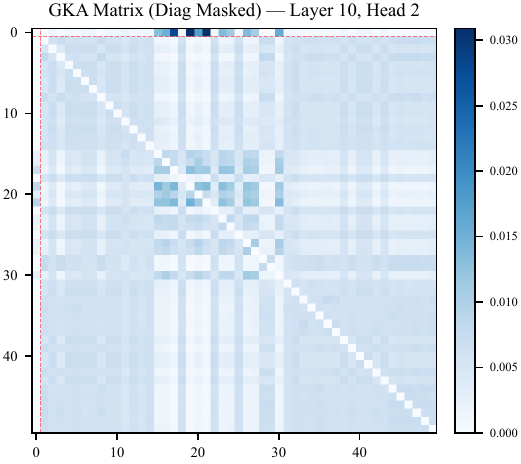}
\caption{}
\label{fig:vlt-attention}
\end{figure*}

\begin{figure*}[tbh]
    \centering
    \includegraphics[width=1\linewidth]{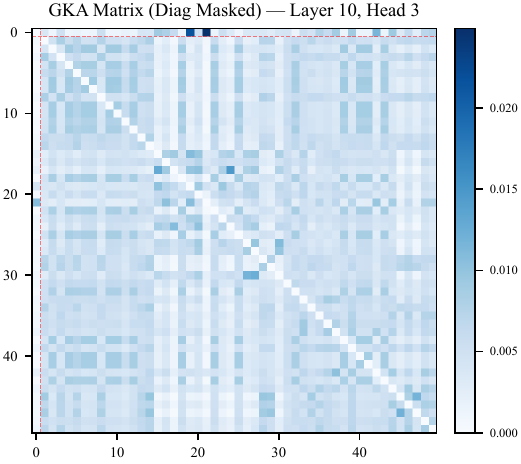}
\caption{}
\label{fig:vlt-attention}
\end{figure*}

\begin{figure*}[tbh]
    \centering
    \includegraphics[width=01\linewidth]{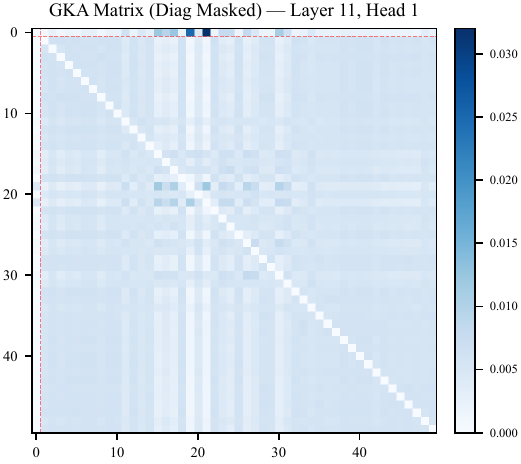}
\caption{}
\label{fig:vlt-attention}
\end{figure*}

\begin{figure*}[tbh]
    \centering
    \includegraphics[width=1\linewidth]{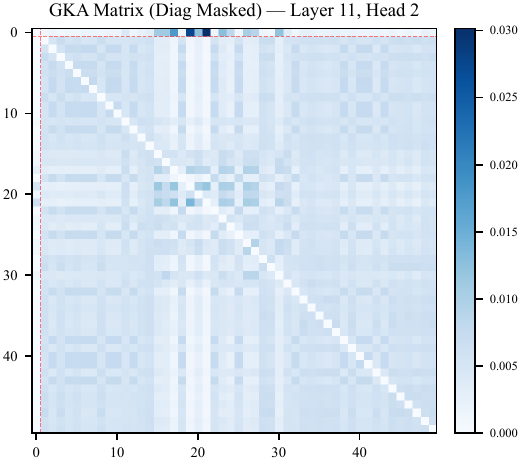}
\caption{}
\label{fig:vlt-attention}
\end{figure*}

\begin{figure*}[tbh]
    \centering
    \includegraphics[width=1\linewidth]{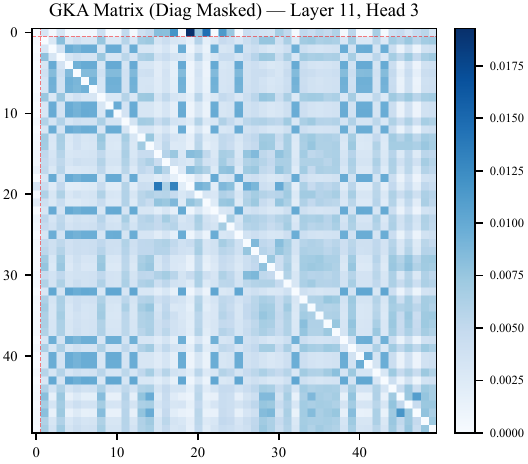}
\caption{}
\label{fig:vlt-attention}
\end{figure*}

\begin{figure*}[tbh]
    \centering
    \includegraphics[width=01\linewidth]{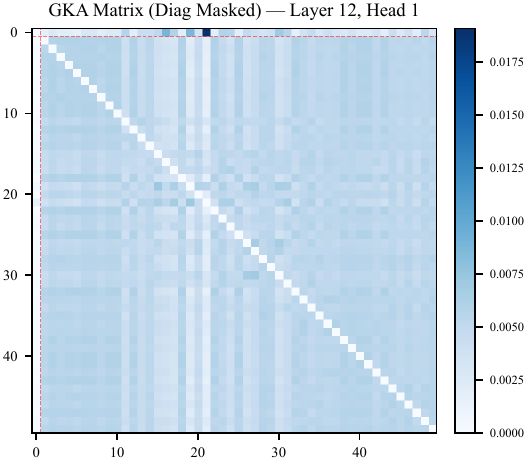}
\caption{}
\label{fig:vlt-attention}
\end{figure*}

\begin{figure*}[tbh]
    \centering
    \includegraphics[width=1\linewidth]{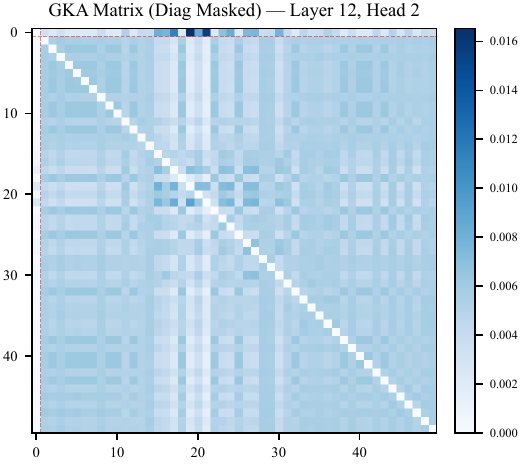}
\caption{}
\label{fig:vlt-attention}
\end{figure*}

\begin{figure*}[tbh]
    \centering
    \includegraphics[width=1\linewidth]{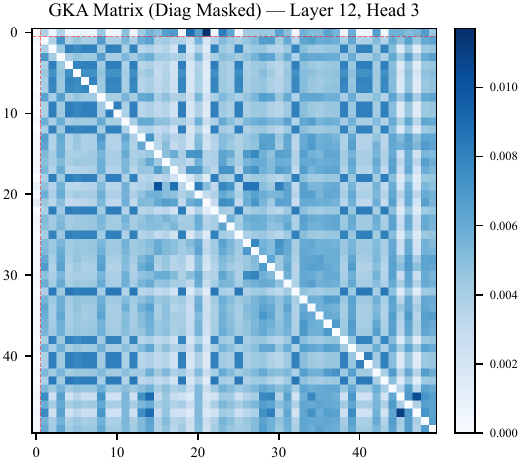}
\caption{}
\label{fig:vlt-attention}
\end{figure*}

\clearpage
\subsection{Implementation Details}
\label{sec:supp_vis_implementation}

All attention maps are extracted via forward hooks registered on each \\ \texttt{GaussianKernelAttention} module, recomputing the kernel from the intermediate features to ensure exact correspondence with the trained model's behavior. The extraction is performed in a single forward pass with no gradient computation. Spatial upscaling from the $14 \times 14$ patch grid to the $224 \times 224$ image resolution uses bicubic interpolation.

\end{document}